\documentclass[runningheads]{llncs}

\usepackage{eccv}

\usepackage{eccvabbrv}

\usepackage{graphicx}
\usepackage{booktabs}
\usepackage{ulem}
\usepackage{multirow}
\usepackage{makecell}
\usepackage{stackengine}
\usepackage{marvosym}

\AtBeginEnvironment{tabular}{\normalsize}

\usepackage[accsupp]{axessibility}  

\usepackage{hyperref}

\usepackage{orcidlink}
\usepackage{siunitx}
\begin{document}


\title{Egocentric World Model for Photorealistic Hand-Object Interaction Synthesis}
\titlerunning{EgoHOI}

\author{
Dayou Li\inst{1}\textsuperscript{*},
Lulin Liu\inst{1,2}\textsuperscript{*},
Bangya Liu\inst{3},
Shijie Zhou\inst{4},
Jiu Feng\inst{5},
Ziqi Lu\inst{6},
Minghui Zheng\inst{1},
Chenyu You\inst{7},
Zhiwen Fan\inst{1}\textsuperscript{\Letter}
}
\authorrunning{D. Li, L. Liu et al.}

\institute{
$^{1}$Texas A\&M University \quad
$^{2}$University of Minnesota \\
$^{3}$University of Wisconsin--Madison \quad
$^{4}$University of California, Los Angeles \\
$^{5}$University of Texas at Austin \quad
$^{6}$Amazon \\
$^{7}$State University of New York at Stony Brook
}

\maketitle

\begingroup
\renewcommand\thefootnote{*}
\footnotetext{Equal contribution.}
\endgroup

\begingroup
\renewcommand\thefootnote{\textdagger}
\footnotetext{Project page: \url{https://egohoi.github.io/}}
\endgroup

\vspace{-2mm}
\begin{abstract}

To serve as a scalable data source for embodied AI, world models should act as true simulators that infer interaction dynamics strictly from user actions, rather than mere conditional video generators relying on privileged future object states. In this context, egocentric Human–Object Interaction (HOI) world models are critical for predicting physically grounded first-person rollouts. However, building such models is profoundly challenging due to rapid head motions, severe occlusions, and high-DoF hand articulations that abruptly alter contact topologies. Consequently, existing approaches often circumvent these physics challenges by resorting to conditional video generation with access to known future object trajectories. We introduce EgoHOI, an egocentric HOI world model that breaks away from this shortcut to simulate photorealistic, contact-consistent interactions from action signals alone. To ensure physical accuracy without future-state inputs, EgoHOI distills geometric and kinematic priors from 3D estimates into physics-informed embeddings. These embeddings regularize the egocentric rollouts toward physically valid dynamics. Experiments on the HOT3D dataset demonstrate consistent gains over strong baselines, and ablations validate the effectiveness of our physics-informed design.

\keywords{Egocentric vision \and World model \and Hand-object Interaction}
\end{abstract}

\vspace{-4mm}

\begin{figure*}[t]
  \centering
  \includegraphics[width=0.9\linewidth]{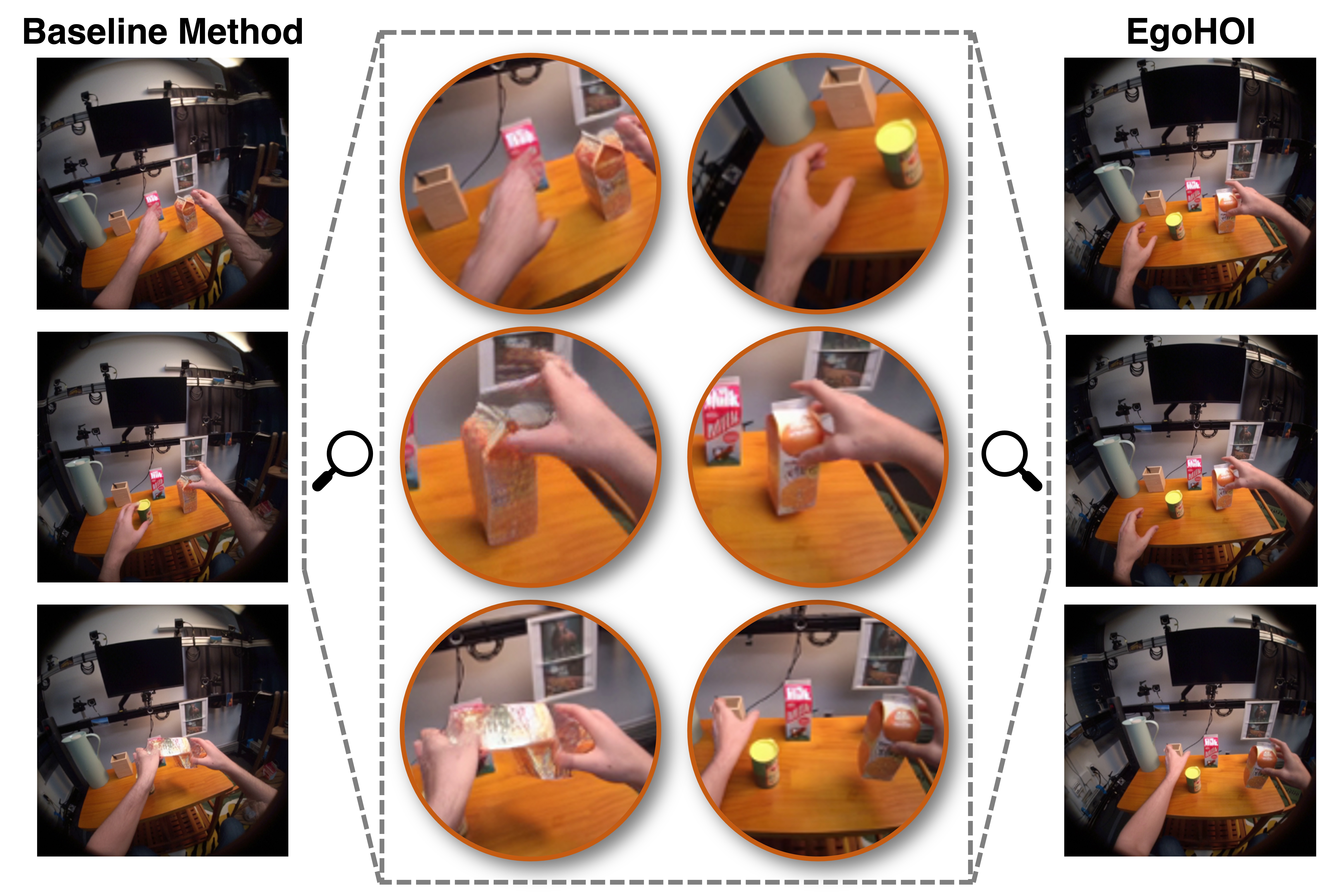} 
  \caption{\textbf{Qualitative comparison between a baseline world model and EgoHOI.} Starting from the same first frame, EgoHOI integrates physics-informed embeddings to model hand–object interaction dynamics, improving ego-motion consistency, kinematic fidelity, and object integrity over time. Zoom-ins highlight clearer contact details and reduced drift.
  }
  \label{fig:teaser}
\end{figure*}

\section{Introduction}
\label{sec:intro}

Egocentric Human-Object Interaction (HOI) world models aim to simulate physically grounded interaction dynamics from a first-person perspective driven by user controls. The resulting predictive rollouts are increasingly important beyond visual synthesis; they can serve as scalable training and evaluation data for a wide range of downstream tasks, including dexterous manipulation \cite{gavryushin2025maple,zheng2026egoscale,luo2026being}, Vision-Language-Actions Models (VLAs) \cite{yang2025egovla,yoshida2025developing,dai2026conla}, spatiotemporal reasoning \cite{zhou2025vlm4d}, physical intelligence \cite{lin2025physbrain}, and data augmentation \cite{lepert2025masquerade}. Despite this potential, the inherent complexity of egocentric HOI poses major challenges. High-DoF hand articulation rapidly changes contact topologies and object states, while severe occlusions and small head motions make the interaction highly ambiguous. The difficulty of this domain is also evident in recent spatiotemporal representation research: as demonstrated by VLM4D \cite{zhou2025vlm4d}, even state-of-the-art VLMs still struggle to reliably understand and reason about the complex dynamics in egocentric HOI videos. If interpreting these interactions is already challenging, building an egocentric HOI world model is even harder. Such a model must go beyond passive observation and actively generate future frames that satisfy user-specified actions, physical laws, and dynamical consistency, producing reliable physical rollouts rather than only visual plausibility.

Despite recent progress in HOI video generation, a reliable egocentric HOI world model remains largely absent. This gap stems from two fundamental limitations in existing pipelines. First, current HOI video generation~\cite{Pang_2025_CVPR, zhang2025vhoi, shen2025idit} is predominantly studied in static, exocentric views. These third-person perspectives do not directly support first-person rollouts with dynamic interactions, which are essential in embodied AI and robotics. Second, many existing approaches~\cite{huang2025hunyuanvideo, jiang2024record,fan2025re,liu2025byteloomweavinggeometryconsistenthumanobject} condition heavily on privileged future object information, such as ground-truth trajectories and waypoints. By explicitly defining how the object should move in the future, these methods largely bypass the core challenge of contact-driven dynamics reasoning. This turns the task into conditional video generation, which conflicts with the formulation of a world model that must infer interaction dynamics and future states strictly from current observations and user-specified actions. Ideally, the model should predict future hand-object interaction purely from action signals, rather than replaying privileged future object states. Because this removes future-state shortcuts, simply scaling RGB data is insufficient for physical accuracy. Instead, the model should be augmented with physics-informed embeddings that inject explicit metric and kinematic structure. Fortunately, recent advances in 3D foundation models for geometry~\cite{wang2025vggt,keetha2025mapanything,dust3r_cvpr24} and hand pose~\cite{pavlakos2024reconstructing} now make it increasingly practical to extract these crucial metric and kinematic priors directly from unstructured videos, opening opportunities to model high-DoF interaction through explicit 3D hand kinematics, camera motion, and contact-aware geometry.

In this paper, we present EgoHOI, an egocentric HOI world model for photorealistic hand-object interaction simulation that integrates physics-informed embeddings into the generative rollout process. EgoHOI uses interpretable estimates from scene, hand, and geometry models, and converts them into embeddings that provide complementary physical structure for learning egocentric dynamics, including hand kinematics, metric-scale ego motion, and object integrity. These embeddings are incorporated through lightweight adapters to preserve the pretrained generative prior while imposing the metric grounding, temporal consistency, and contact-aware dynamics. Unlike earlier trials utilizing access to privileged future states, EgoHOI enables causal simulation of physically plausible hand-object interactions while retaining the generative diversity and reconstruction quality trained from massive data. We demonstrate an example in~\cref{fig:teaser} to show the resulting simulated rollouts, and summarize our contributions as follows:
\begin{itemize}
    \item We introduce EgoHOI, an egocentric HOI world model that simulates first-person hand–object interaction dynamics under user-specified actions and controls, without privileged future object states.
    \item We introduce physics-informed embeddings derived from 3D geometry and kinematic priors to regularize egocentric HOI rollouts toward physically consistent interaction dynamics.
    \item We show EgoHOI better aligns rollouts with user actions and improves hand–object interaction realism, with comprehensive metrics and ablations validating our physics-informed embeddings.
\end{itemize}
\section{Related Works}
\label{sec:related_work}
\noindent\textbf{Egocentric World Model \& Video Generation.}
Egocentric World model aims to simulate the physical world by predicting future visual observations in a first-person viewpoint. 
Recently, there has been progress in the general world models~\cite{chen2025planningreasoningusingvision,routray2025vipravideopredictionrobot}, which usually take in detailed language descriptions as control signals to generate future rollouts. For instance, Nvidia's Cosmos~\cite{nvidia2025cosmosworldfoundationmodel}, which was trained on 100M language-video pairs, can capture diverse real-world physics and be fine-tuned for specific tasks. However, most of these models operate primarily in image space, which makes physically plausible rollouts under high degrees of freedom still challenging in practice.
Unlike general world models, egocentric world models often focus on more challenging tasks. In egocentric settings, viewpoints constantly change, while hand-object interactions frequently influence the dynamic evolution of the scene, making it significantly more difficult for world models to simulate~\cite{grauman2022ego4d,huang2025understanding,bai2025whole}. PlayerOne~\cite{tu2025playerone} drives a first-person simulator with whole-body motion to maintain consistent visual rendering across scenes, while GEM~\cite{hassan2025gem} jointly generates RGB and depth streams conditioned on ego-trajectory and object identity features. 
Beyond PlayerOne and GEM, many existing efforts focus on designing view-transformation or hand-trajectory controls for first-person generation, rather than grounding rollouts in reconstruction-based 3D priors or causal hand–object dynamics. EgoExo-Gen~\cite{xuegoexo} is one example that generates a first-person video by watching a third-person view of the same scene. Zhang et al.~\cite{anonymous2025egocentric} first predict a coarse hand trajectory and then condition a latent diffusion model to generate future frames.

Overall, prior egocentric world models focus on plausible first-person rendering, but hand–object interaction is less emphasized, and maintaining physically consistent interaction dynamics remains difficult.

\noindent\textbf{Hand-Object Interactions Synthesis.}
Hand–object interaction (HOI) synthesis targets interactions that are visually realistic and physically plausible by jointly modeling appearance, relative geometry, and contact semantics~\cite{li2025latenthoi,christen2024diffh2o,diller2024cg,li2024controllable,liu2024geneoh,ghosh2023imos,lee2024interhandgen}.
In the egocentric setting, MEgoHand~\cite{zhou2025megohand} synthesizes hand–object motion from first-person RGB together with text and an initial hand pose.
Beyond egocentric inputs, a dominant trend in general HOI generation is reference-conditioned interaction alignment, where models leverage reference signals such as target poses, relative transforms, or motion trajectories to improve controllability and contact quality. These include image-level refinement of hand–object alignment~\cite{jiang2024record}, geometry and relation aware motion generation with diffusion~\cite{xue2025guiding}, multimodal video generation guided by an HOI adapter~\cite{huang2025hunyuanvideo}, open world synthesis with affordance guidance and physics refinement~\cite{zhang2025openhoi}, and tokenized sequence models that translate between language and long 3D HOI sequences~\cite{Huang_2025_CVPR}.

\begin{figure*}[t]
  \centering
  \includegraphics[width=\textwidth]{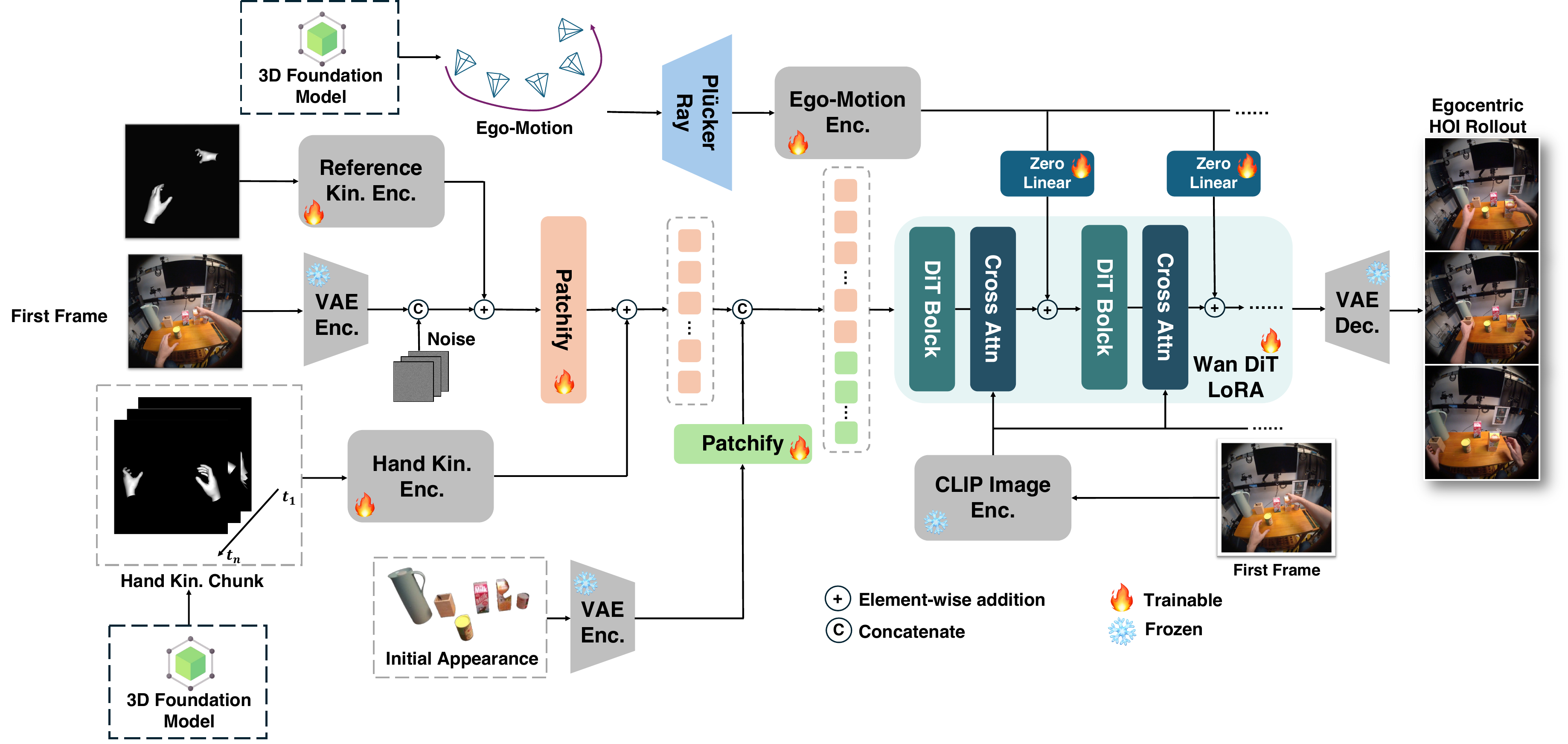} 
  \caption{\textbf{Overview of EgoHOI pipeline. }
 We formulate EgoHOI as an egocentric world model that represents frames with a latent internal state and predicts action-driven transitions with a DiT backbone. Physics-informed embeddings distilled from reconstruction-based 3D priors, together with the first-frame object appearance, are integrated into the latent dynamics via lightweight adapters, enabling realistic hand–object interactions, geometry-consistent ego-motion, and stable object identity under viewpoint changes.
  }
  \label{fig:pipeline}
\end{figure*}
%

\section{Method}

\subsection{Preliminary}
\subsubsection{World Model Formulation}
\label{sec:world_model_formulation}

A world model $\mathbf{f_\theta}$ captures action-driven internal state dynamics and enables rollouts by iteratively predicting how the state evolves under actions.
Let $x_t \in \mathcal{X}$ denote the internal state and let $a_t$ denote the action applied at time $t$.
The action-driven transition dynamics are:
\begin{equation}
x_{t+1} = \mathbf{f_\theta}(x_t, a_t),
\label{eq:wm_dynamics}
\end{equation}
and the observation function $g_\phi$ maps the internal state to the observation space:
\begin{equation}
o_t \sim g_\phi(x_t),
\label{eq:wm_obs}
\end{equation}
where $o_t$ denotes the observable signal.
Given an initial observation $o_1$, an encoder produces the initial internal state $x_1 = e_\phi(o_1)$.
A rollout over horizon $L$ is obtained by recursively applying $\mathbf{f_\theta}$ under an action sequence $\{a_t\}_{t=1}^{L}$ and mapping the resulting latent states to predicted observations via $g_\phi$.

In this view, a world model serves as an action-driven simulator: actions are treated as interventions that induce state transitions, and rollouts should predict the causal effects of actions on the world state, instead of only reconstructing the agent’s kinematic trajectory. In EgoHOI’s setting, we instantiate the transition model $\mathbf{f_\theta}$ with a DiT backbone to realize action-driven evolution of the model’s internal state in a VAE latent space; $e_\phi$ and $g_\phi$ are the VAE encoder and decoder that map video frames to and from this latent representation.

\subsubsection{Diffusion Process}
Video generative models~\cite{ho2022video,blattmann2023stable} generates samples by iteratively denoising random noise through a learned reverse diffusion process. 
Formally, we consider a latent representation $z_{0}$ of an initial clean data sample obtained from a variational autoencoder (VAE) that encodes the video into its latent space. A forward (noising) process $q$ gradually adds Gaussian noise to $z_{0}$ over $T$ time steps, producing a sequence $z_{1}, z_{2}, \dots, z_{T}$. At each step, noise is added according to a variance schedule ${\beta}_{t=1}^T$. A common formulation is:
\begin{equation}
q(z_t|z_{t-1})
=\mathcal{N}(z_t;\sqrt{1-\beta_t}\,z_{t-1},\,\beta_t\mathbf{I}),~
t=1,...,T
\end{equation}
with $\beta_t \in (0,1)$ controlling the noise level. As $t$ increases, the latent $z_t$ becomes increasingly noisy; in the limit $z_{T}$ approaches a standard Gaussian noise. Equivalently, one can express $z_t$ in closed form as a noisy mixture of the original latent and noise:
\begin{equation}
    z_t = \sqrt{\bar{\alpha}_t}\, z_0 + \sqrt{1 - \bar{\alpha}_t}\, \varepsilon
\end{equation}
where $\alpha_t := 1-\beta_t$, and $\bar\alpha_t := \prod_{s=1}^{t}\alpha_s$ is the accumulated noise attenuation. Here $\epsilon \sim \mathcal{N}(0,\mathbf{I})$ represents standard Gaussian noise. The reverse process learns to invert this degradation by removing the noise from $z_t$ to reconstruct the original latent representation. The denoiser is trained to predict the noise added to the latent, with respect to the loss function:
\begin{equation}
    \mathcal{L}
= 
\mathbb{E}_{z_0,\, \varepsilon,\, t}\!\left[
\big\|\, \varepsilon - \varepsilon_\theta(z_t,\, t)\, \big\|_2^2
\right]
\end{equation}
By minimizing this objective, the model learns to remove noise and recover the clean latent $z_0$ from a noisy $z_t$.

\subsection{Latent-Space World Modeling for Egocentric HOI}
We model egocentric hand--object interaction (HOI) synthesis as action-driven world modeling in a compact latent state space. Given the first-frame observation $\mathbf{I_1}\in\mathbb{R}^{H\times W\times 3}$, we obtain an initial latent state $z_1$ via the VAE encoder. The action signal at each step $t$ is specified by hand kinematics $\mathbf{H}^t$ and metric head motion $\mathbf{C}^t$, forming an $L$-step action sequence $\{(\mathbf{H}^t,\mathbf{C}^t)\}_{t=1}^{L}$. Our goal is to predict the temporal evolution of latent states under actions and decode them into future first-person observations.
Concretely, the world model defines action-driven latent dynamics that evolve the state over time,
\begin{equation}
z_{t+1} = \mathbf{f_\theta}(z_t,\, \mathbf{H}^t,\, \mathbf{C}^t), \qquad t=1,\dots,L,
\end{equation}
and the predicted observation rollout is obtained by decoding $\{z_t\}$ into $\hat{\mathbf{I}}_{2:L+1}$.

Modeling egocentric HOI dynamics in latent space is challenging due to the coupled effects of high-DoF hand articulation, metric ego motion, and contact-driven object dynamics. We regularize the action-driven latent evolution without relying on privileged future object states. The next section introduces physics-informed embeddings distilled from 3D geometric and kinematic estimates, which promote physically plausible latent dynamics.

\subsection{Physics-Informed Embeddings}
To constrain HOI rollouts without using privileged future object states, we distill 3D geometric and kinematic estimates into compact physics-informed embeddings.
Concretely, we derive three embeddings from complementary sources of structure: (i) hand kinematics that describe high-DoF articulation, (ii) metric ego motion that encodes the viewpoint trajectory, and (iii) an object-entity anchor tied to the first frame to stabilize object-centric appearance under interaction.
These embeddings are injected into the latent dynamics to provide per-step geometric and kinematic context, thereby guiding state evolution toward physically plausible dynamics.

\subsubsection{Hand Kinematic Embeddings (HKE).}
The goal of HKE is to capture dense hand motion structure from reconstructed 3D hand estimates, which is essential for contact-consistent HOI rollouts under high-DoF articulation. 
Rather than representing actions as sparse coordinate vectors, we use explicit per-frame hand renders obtained from reconstructed meshes as a dense 3D representation. We stack a sequence of hand render frames $\{H_t\}_{t=1}^{L}$ and encode them with a temporal 3D convolutional stack with SiLU activations, which gradually increases the channel dimension and maps the hand stream to the shared the $5120$-dimension latent space.
To stabilize hand identity and reduce temporal drift, we add a reference-hand branch that encodes the first-frame hand pose $\mathbf{P}^1$ into a compact feature $\mathbf{R}_{\mathrm{ref}}\in\mathbb{R}^{H''\times W''\times 20}$ via a six-layer 2D convolutional pyramid ($3\times 3$ Conv + SiLU). Together, the temporal 3D Conv stack and the reference encoder produce hand kinematic embeddings that preserve fine-grained articulation structure and support physically plausible hand evolution during rollouts.

\subsubsection{Ego-Motion Embeddings (EME).}
Ego head motion is a dominant factor in first-person rollouts, where viewpoint changes couple tightly with hand motion and occlusions. EME encodes metric head trajectory estimates into embeddings aligned with the latent space, enabling geometry-consistent viewpoint evolution over time. Given a head pose sequence $\{\mathbf{C}^t\}_{t=1}^{L}$ with intrinsics $\mathbf{K}\in\mathbb{R}^{3\times 3}$ and extrinsics $\mathbf{E}=[\mathbf{R}:\mathbf{t}]\in\mathbb{R}^{3\times 4}$ (relative to the first frame), we adopt Pl\"ucker embeddings to represent the per-pixel ray field induced by the head pose. For each pixel $(u,v)$, its embedding is $\mathbf{p}_{u,v}=(\mathbf{o}\times \mathbf{d}_{u,v},\, \mathbf{d}_{u,v})\in\mathbb{R}^{6}$, where $\mathbf{o}$ is the head-frame optical center and $\mathbf{d}_{u,v}$ is the world-direction ray computed as:
\begin{equation}
\mathbf{d}_{u,v}=\mathbf{R}\,\mathbf{K}^{-1}\begin{bmatrix}u\\v\\1\end{bmatrix}+\mathbf{t}.
\end{equation}
This ray-based representation is invariant to the choice of scene origin and preserves metric structure under head motion. By representing ego motion as an image-aligned ray field, it avoids collapsing geometry into a low-dimensional pose vector and instead imposes spatially resolved constraints on viewpoint evolution. We stack per-frame Pl\"ucker tensors into a trajectory representation, process it with causal 3D convolutions and hybrid spatiotemporal blocks that combine 2D residual convolutions with temporal self-attention, and finally use a projection head and patchifier to map features into the latent grid, yielding ego-motion embeddings that regularize long-horizon viewpoint dynamics.

\subsubsection{Object Entity Embeddings (OEE).}
Egocentric rollouts can suffer from object entity drift, which often appears as instability in color, texture, and fine details under rapid ego motion and partial occlusions.
To stabilize object entities over time, OEE anchors each object to the first frame, providing a persistent reference that supports coherent HOI evolution across the rollout.
Concretely, we use the first-frame object segmentation image, obtained from an off-the-shelf segmentation model, to isolate object-centric evidence and reduce background interference under viewpoint changes.
The segmentation is encoded by the frozen video VAE encoder into a latent grid, and a 3D convolutional patchifier converts it into tokens aligned with the main latent stream.
Operating in the same latent parameterization as the backbone, OEE yields features that mitigate drift and preserve object-centric details throughout the rollout.


\subsection{Latent Space Integration}
We incorporate physics-informed embeddings into the frozen Wan-DiT in a single token-space forward pass by first aligning each embedding stream to the main token layout and shared token dimension $d$. Let the Wan-VAE latent be $z\in \mathbb{R}^{B \times F \times H \times W \times C}$ and the corresponding main token sequence be $\mathbf{T}_{\text{main}} \in \mathbb{R}^{B \times N \times d}$. Each embedding tensor is projected and patchified into an aligned token sequence compatible with $\mathbf{T}_{\text{main}}$, enabling Wan-DiT to leverage physics-informed structure for latent evolution without modifying the original backbone architecture.

\subsubsection{Action-Driven Token.}
The hand encoder produces hand tokens $\mathbf{T}^{\text{HKE}}$ in $\mathbb{R}^{B \times N \times d}$, which are fused into the main token stream via a gated residual update:
\begin{equation}
\mathbf{X}_0 = \mathbf{T}_{\text{main}} + \gamma_h \mathbf{T}^{\text{HKE}}
\end{equation}
where $\gamma_h$ is a learnable gate. In parallel, the reference-hand branch produces a first-frame hand feature that is added to the first-frame latent and concatenated with the noise latent before patchification, tying the initial hand configuration to the latent state space.
The ego-motion adapter outputs geometry tokens $\mathbf{T}^{\text{EME}} \in \mathbb{R}^{B \times N \times d}$. For compatibility with the extended sequence length introduced below, we zero-pad $\mathbf{T}^{\text{EME}}$ to obtain  with the same length as the active token sequence. In the first $D$ DiT blocks, we apply per-block zero-initialized linear adapters $\mathbf{U}_l : \mathbb{R}^d \rightarrow \mathbb{R}^d$ to generate residuals:
\begin{equation}
\Delta \mathbf{X}_l = \mathbf{U}_l\!\left(\mathbf{T}^{\text{EME}}\right)
\end{equation}
which are added to the block input. This construction regularizes viewpoint evolution under metric head motion while preserving backbone stability at initialization.

\subsubsection{Object-Entity Token.}
A object patchifier converts first-frame object entity features into tokens
$T^{\mathrm{OEE}} \in \mathbb{R}^{B \times N \times d}$.
We assign shifted RoPE to spatially anchor these tokens as an independent reference stream, then append them to form the extended sequence
\begin{equation}
T_{\mathrm{all}} = [X_0;\, T^{\mathrm{OEE}}]
\end{equation}
Within each DiT block, the object-entity tokens participate as KV-only context during self-attention; only the main tokens are decoded, which provides a persistent entity reference that reduces drift in long rollouts.

\subsubsection{First-Frame Anchor}
The first frame is encoded by the Wan VAE encoder as $\mathbf{Y}$, and the reference-hand feature is added to $\mathbf{Y}$ to couple initial kinematics and appearance. Before patchification, we concatenate $\mathbf{Y}$ with the noise latent along the channel dimension. In parallel, the first frame is processed by the CLIP image encoder~\cite{radford2021clip} to obtain global image embeddings; these embeddings are linearly projected and provided as fixed keys and values to the cross-attention layers of Wan-DiT, encouraging preservation of overall scene structure and appearance consistency during denoising.

\begin{figure*}[t]
  \centering
  \includegraphics[width=0.65\textwidth]{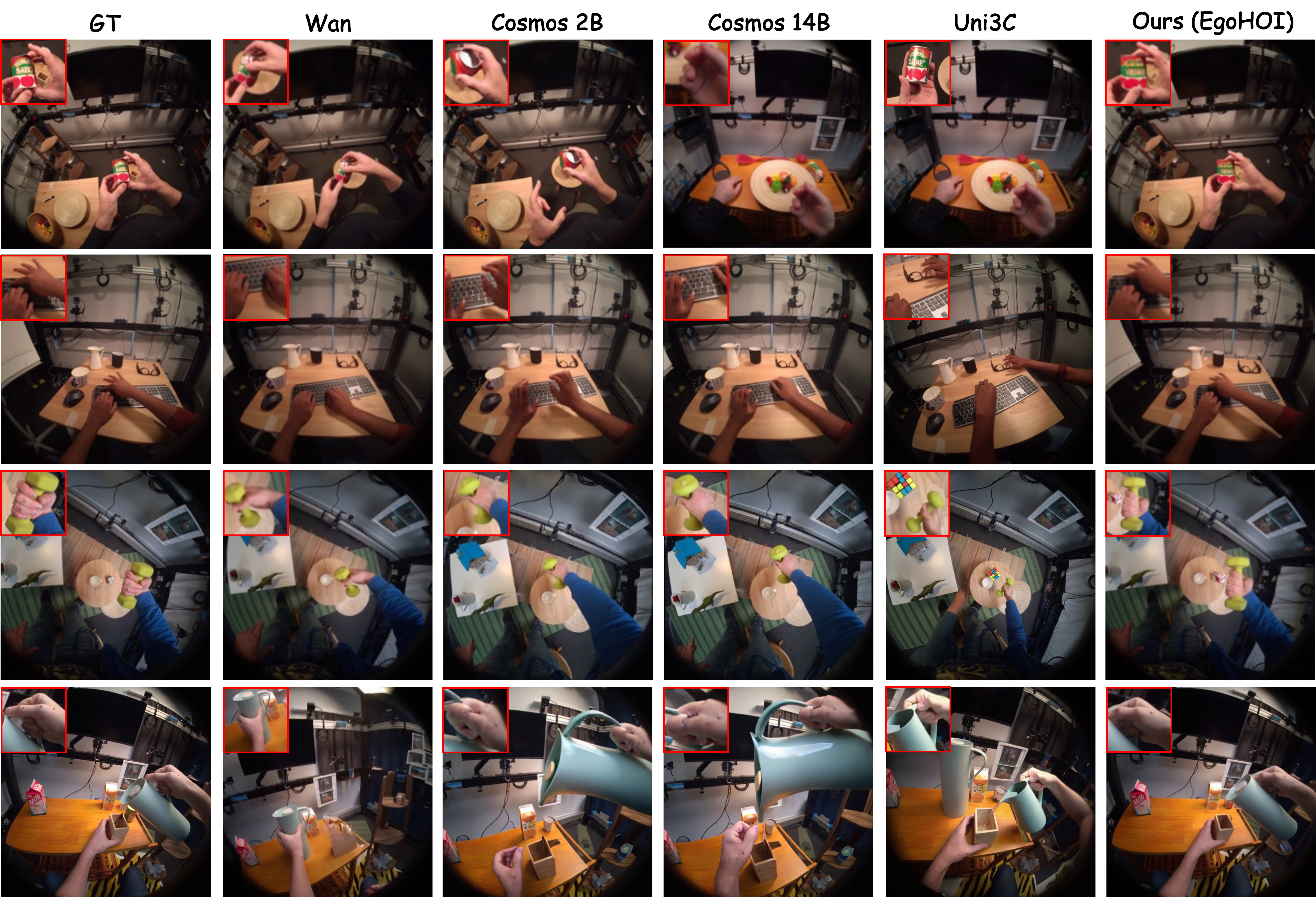} 
  \vspace{-3mm}
  \caption{\textbf{Qualitative comparison with baselines.}
 Columns from left to right show the ground truth (GT), Wan, Cosmos-2B, Cosmos-14B, Uni3C, and our EgoHOI model. All methods receive the same first frame as input; compared with the baselines, our model better preserves hand and object geometry, maintains object identity, and produces more stable interaction dynamics over time. See the zooming boxes for comparison of the fine-grained details.
  }
  \label{fig:sota2}
\end{figure*}
\section{Experiment}

\begin{table*}[t]
\vspace{-3mm}
\caption{\textbf{Comparison with baselines.} We compare EgoHOI against a Wan backbone, Cosmos world models (2B, 14B) and Uni3C on frame prediction (PSNR/SSIM/LPIPS/Object-CLIP), ego-motion consistency (ATE/RRE/RPE), and kinematic fidelity (MR/MPJPE/RMSE). $\uparrow$ higher is better, $\downarrow$ lower is better.}
\label{tab:sota}
\begin{center}
\resizebox{\textwidth}{!}{
\begin{tabular}{lcccccccccc}
\toprule
\multirow{2}{*}{\bf Method}  & \multicolumn{4}{c}{Frame Prediction} & \multicolumn{3}{c}{Ego-Motion Consistency} & \multicolumn{3}{c}{Kinematic Fidelity} \\
\cmidrule(r){2-5}\cmidrule(r){6-8}\cmidrule(r){9-11}
& \multicolumn{1}{c}{PSNR $\uparrow$}
& \multicolumn{1}{c}{SSIM $\uparrow$}
& \multicolumn{1}{c}{LPIPS $\downarrow$}
& \multicolumn{1}{c}{Object-CLIP $\uparrow$}
& \multicolumn{1}{c}{ATE $\downarrow$}
& \multicolumn{1}{c}{RRE $\downarrow$}
& \multicolumn{1}{c}{RPE $\downarrow$}
& \multicolumn{1}{c}{MR $\downarrow$}
& \multicolumn{1}{c}{MPJPE $\downarrow$}
& \multicolumn{1}{c}{RMSE $\downarrow$} \\
\cmidrule(r){1-1}\cmidrule(r){2-5}\cmidrule(r){6-8}\cmidrule(r){9-11}
Wan~\cite{wang2025wan}  & 14.78  & 0.46 & 0.52 & 0.86 & 0.124 & 10.687 & 0.036 & 19.67\% & 0.049 & 0.083 \\
Cosmos 2B~\cite{nvidia2025cosmosworldfoundationmodel} & 15.49 & 0.56 & 0.41 & 0.87 & 0.116 & 9.515 & 0.026 & 16.23\% & 0.044 & 0.078 \\
Cosmos 14B~\cite{nvidia2025cosmosworldfoundationmodel} & 15.89 & 0.59 & 0.38 & 0.88 & 0.112 & 9.046 & 0.022 & 14.61\% & 0.041 & 0.074 \\
Uni3C~\cite{cao2025uni3c} & 14.21 & 0.50 & 0.55 & 0.82 & 0.168 & 14.209 & 0.026 & 20.50\% & 0.073 & 0.089 \\
\midrule
Ours (EgoHOI) & \textbf{21.05} & \textbf{0.65} & \textbf{0.27} & \textbf{0.92} & \textbf{0.084} & \textbf{5.192} & \textbf{0.021} & \textbf{5.84\%} & \textbf{0.014} & \textbf{0.044} \\
\bottomrule
\end{tabular}
}
\end{center}
\end{table*}

\subsection{Experiment Setup}
\subsubsection{Implementation Details}
We choose Wan 2.1 14B~\cite{wang2025wan} as our model backbone. We start from the official checkpoint with LoRA rank set to 128 and LoRA alpha set to 128. The inference step and the learning rate are set to 50 and $1\times10^{-5}$. We use the Adam optimizer with BF16 precision. The cfg is set to 1.0. We train our model for 8000 steps on 16 NVIDIA H100 GPUs with a batch size of 1 and a sample resolution of $480\times480$. The training process lasts for approximately 1 day. 
During training and inference, our frame length is set to 81. After training, our model can achieve 16 FPS when simulating human-object interaction.

\subsubsection{Benchmark}
Since there is no publicly available benchmark for world modeling in egocentric hand–object interaction, we construct an evaluation benchmark from HOT3D. We select 100 clips from scenes that are excluded from the training set. Each clip contains 150 frames and provides hand mesh renders and camera trajectories. Although our model operates on 81-frame rollouts, we make full use of each 150-frame clip by extracting multiple 81-frame sub-clips with a sliding-window scheme, where the window length is fixed to 81 frames and is shifted along the clip to form evaluation samples.

\subsubsection{Metrics}
We evaluate the model from three complementary aspects: visual prediction, ego-trajectory consistency, and kinematic fidelity. Visual prediction is measured by PSNR, SSIM~\cite{wang2004ssim}, LPIPS~\cite{zhang2018lpips}, and Object-CLIP~\cite{xu2024anchorcrafter}. We additionally report VBench~\cite{zheng2025vbench} scores to assess perceptual and temporal quality, including Subject Consistency, Background Consistency, Motion Smoothness, Dynamic Degree, Aesthetic Quality, and Imaging Quality. Ego-motion consistency is quantified by ATE, RPE, and RRE between the estimated and ground-truth camera trajectories. Kinematic accuracy is computed with HaMeR~\cite{pavlakos2024reconstructing} on both ground-truth and generated results, reporting Missing Ratio (MR), MPJPE, and RMSE on hand segmentation images.
Details are in the supplementary material.

\subsection{Comparison with Baselines}
We compare EgoHOI with three tiers of baselines for egocentric HOI rollouts. Wan~\cite{wang2025wan} serves as a backbone-level video generation baseline. Cosmos 2B and 14B~\cite{nvidia2025cosmosworldfoundationmodel} are general-purpose world models at two parameter scales. We also include Uni3C~\cite{cao2025uni3c}, which offers stronger camera and human motion control. Wan and Cosmos are fine-tuned from released checkpoints using the same data and protocol as EgoHOI, and all methods are evaluated on the same datasets, metrics, and pipeline. Several recent egocentric world models~\cite{tu2025playerone,xie2026generatedrealityhumancentricworld} are omitted because their implementations are not publicly available at submission time, which prevents a reproducible evaluation under our protocol.

Table~\ref{tab:sota} reports frame prediction, ego-motion consistency, and kinematic fidelity. Scaling from Wan to Cosmos yields modest gains in frame quality, but motion-related performance remains limited. For example, Cosmos 14B reaches ATE 0.112 and MR 14.61\%. Uni3C improves motion control, yet it degrades under rollout evaluation, with the largest ego-motion errors and weakest kinematic fidelity (ATE 0.168, RRE 14.209, MR 20.50\%), suggesting that motion control alone does not recover contact-driven interaction dynamics. In contrast, EgoHOI performs best across all metric groups. It increases PSNR to 21.05 while substantially reducing ego-motion and hand kinematic errors, indicating that our physics-informed embedding extraction and injection improve interaction dynamics and temporal physical consistency, rather than only per-frame quality. Additional object-level comparisons are provided in the supplementary material, including OPE and OOE from our ablation study.

For VBench in \cref{tab:vbench2}, EgoHOI achieves the highest subject consistency (95.51\%), background consistency (94.91\%), aesthetic quality (52.03\%), and imaging quality (64.48\%). Its lower Dynamic Degree (53.29\%) reflects a design choice that prioritizes stable hand--object interactions over highly dynamic camera motion. Qualitatively (\cref{fig:sota2}), Wan exhibits drift; Cosmos produces sharper frames but shows contact-region errors and small-object inconsistencies; Uni3C constrains camera motion but remains unreliable around contact. Overall, changing backbones, scaling model size, or enforcing motion control does not address the core challenge of egocentric HOI world models, whereas EgoHOI regularizes action-driven dynamics with physics-informed embeddings distilled from 3D priors to yield more plausible rollouts. More visualization is included in the supplementary material. 

\begin{table*}[h]
    \vspace{-4mm}
  \caption{\textbf{VBench Evaluation of the Baselines.} Reported scores cover subject consistency, background consistency, motion smoothness, dynamic degree, aesthetic quality, and imaging quality. For the consistency and perceptual quality metrics, higher values correspond to better results, whereas dynamic degree characterizes how dynamic the generated content is rather than its realism.}
  \label{tab:vbench2}
  \centering
  \resizebox{0.8\textwidth}{!}{%
  \begin{tabular}{@{}l|c|c|c|c|c|c@{}}
    \specialrule{1.2pt}{0pt}{0pt}
    \textbf{Method} & \makecell[c]{Subject \\ Consistency} & \makecell[c]{Background \\ Consistency} & \makecell[c]{Motion \\ Smoothness} & \makecell[c]{Dynamic \\ Degree} & \makecell[c]{Aesthetic \\ Quality} & \makecell[c]{Imaging \\ Quality} \\
    \specialrule{1.2pt}{0pt}{0pt}
    Wan~\cite{wang2025wan}        & 91.40\% & 93.14\% & 99.19\% & 64.75\% & 46.57\%  & 58.58\% \\
    Cosmos 2B~\cite{nvidia2025cosmosworldfoundationmodel}   & 89.01\% & 93.00\% & 99.27\% & 87.50\% & 46.88\% & 62.98\% \\
    Cosmos 14B~\cite{nvidia2025cosmosworldfoundationmodel} & 93.21\% & 93.56\% & 99.32\% & \textbf{89.50\%} & 49.83\% & 63.78\% \\
    Uni3C~\cite{cao2025uni3c}      & 91.91\% & 94.29\% & 99.33\% & 49.00\% & 47.03\% & 61.47\% \\
    \specialrule{1.2pt}{0pt}{0pt}
    Ours (EgoHOI)       & \textbf{95.51\%} & \textbf{94.91\%} & \textbf{99.40\%} & 53.29\% & \textbf{52.03\%} & \textbf{64.48\%} \\
    \specialrule{1.2pt}{0pt}{0pt}
  \end{tabular}
  }
  \vspace{-3mm}
\end{table*}

\begin{figure}[t]
  \centering
  \includegraphics[width=0.8\textwidth]{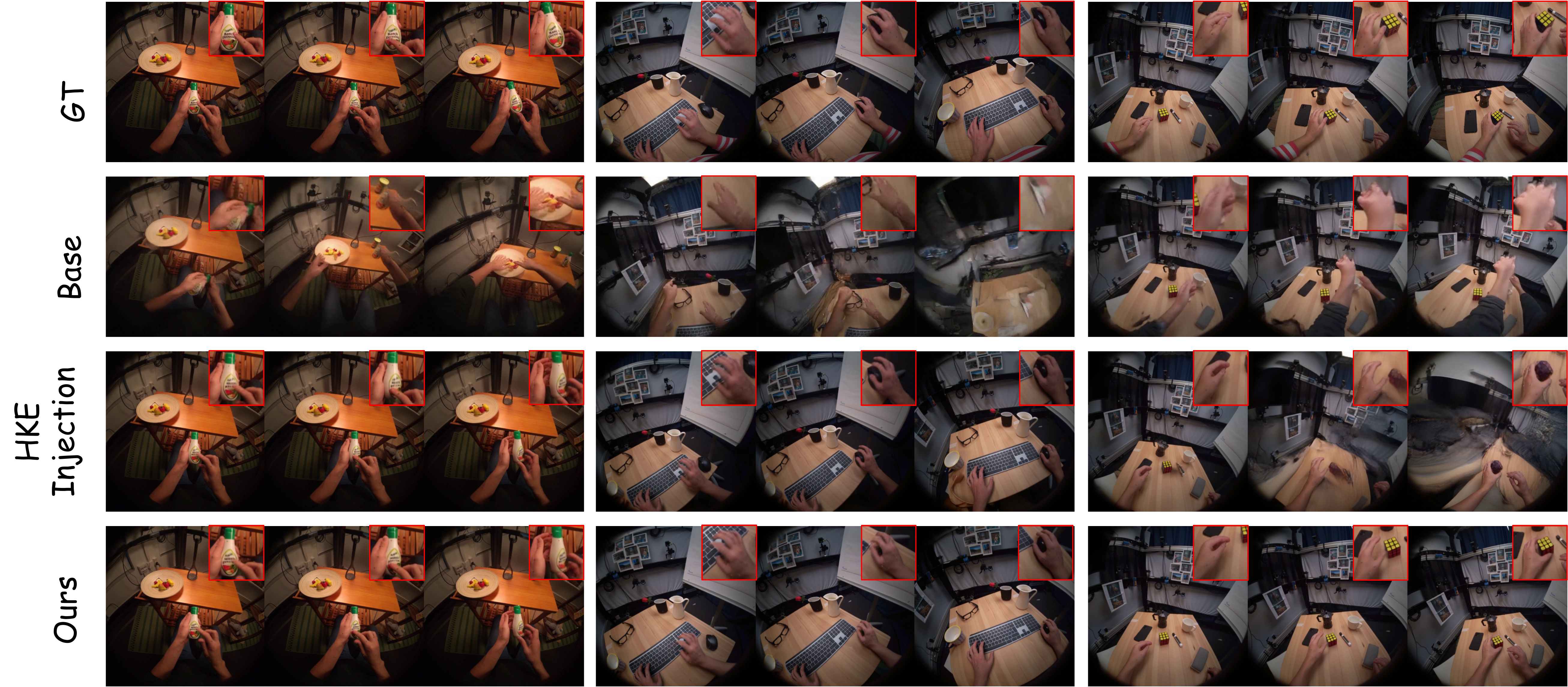} 
  \caption{\textbf{Qualitative results for ablation on kinematic fidelity.}
Rows: GT, Base, HKE Injection, Ours (EgoHOI). HKE improves hand articulation stability and reduces local shape distortion. Zoom-in boxes highlight clearer hand–object contact details. The full model is the most stable, matching MR/MPJPE/RMSE.
  }
  \label{fig:KIN}
\end{figure}
\subsection{Ablation Study}
We provide a detailed ablation analysis of EgoHOI by isolating how each component supports egocentric hand--object rollouts along three aspects: kinematic fidelity, object entity stability, and ego-motion consistency. Across kinematic fidelity (\cref{tab:ablationKIN}), ego-motion consistency (\cref{tab:ablationGEO}), and object integrity (\cref{tab:ablationAPP}), we compare several configurations of EgoHOI. Base uses the Wan 2.1 DiT video backbone without HKE, OEE, or EME. The HKE Injection, OEE Injection, and EME Injection variants each enable a single component on top of this backbone, which isolates its contribution under otherwise identical settings. EgoHOI activates three components jointly and corresponds to the model used in the main experiments. All variants share the same data, training and optimization settings. We measure kinematic fidelity with MR, MPJPE, and RMSE, object entity with Object-CLIP together with VBench Subject Consistency and Imaging Quality, and ego-motion consistency with ATE, RPE, and RRE.

\subsubsection{Study on Kinematic Fidelity}

\cref{tab:ablationKIN} examines the effect of HKE on hand state evolution over egocentric rollouts. Compared with the base variant, adding HKE reduces the MR and MPJPE to 6.47\% and 0.015 from 28.77\% and 0.576, respectively, indicating more reliable hand localization and substantially improved joint accuracy under interaction and partial occlusions. The reduction in RMSE further suggests that the hand region matches the ground truth more closely in both shape and RGB appearance over the sequence. The full model, which combines HKE with OEE and EME, further improves these metrics to 5.84\%, 0.014, and 0.044. Qualitatively (\cref{fig:KIN}), HKE mitigates hand-shape collapse and temporal fluctuations, while the full model yields the most stable hand evolution across time with fewer distortions during frequent hand--object interaction.

\begin{table}[h]
\setlength{\abovecaptionskip}{5pt}
  \caption{\textbf{Ablation on kinematic fidelity.} We ablate HKE Injection and report hand motion fidelity in generated videos using MR, MPJPE, and RMSE; $\downarrow$ indicates lower is better.}
  \label{tab:ablationKIN}
  \centering
  \renewcommand{\arraystretch}{1.15}
  \setlength{\tabcolsep}{4.5pt}
  \resizebox{0.5\linewidth}{!}{ 
    \begin{tabular}{lccc}
      \toprule
      \textbf{Method} 
        & \makecell[c]{MR} $\downarrow$
        & \makecell[c]{MPJPE} $\downarrow$
        & \makecell[c]{RMSE} $\downarrow$ \\
      \midrule
      Ours (base) & 28.77\% & 0.576 & 0.089 \\
      HKE Injection & 6.47\% & 0.015 & 0.044 \\
      \midrule
      \textbf{Ours (EgoHOI)} & \textbf{5.84\%} & \textbf{0.014} & \textbf{0.044} \\
      \bottomrule
    \end{tabular}
  }
\end{table}

\subsubsection{Study on Ego-Motion Consistency}

\begin{figure}[h]
  \centering
  \includegraphics[width=0.8\textwidth]{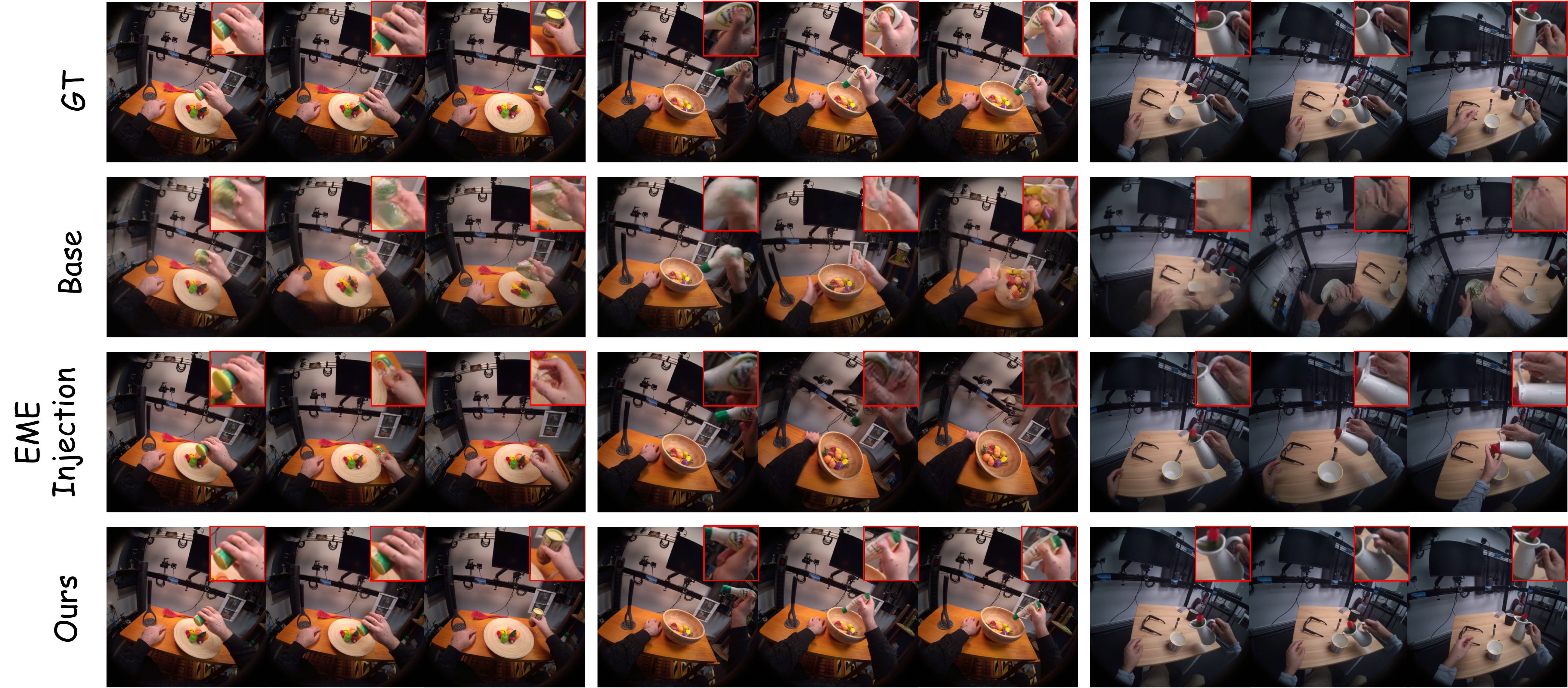} 
  \caption{\textbf{Qualitative results for ablation of ego-motion consistency }
Rows: GT, Base, EME Injection, Ours (EgoHOI). EME mitigates long-horizon ego-motion drift and improves viewpoint stability. Zoom-ins highlight cleaner details under viewpoint changes. The full model best matches ATE/RRE/RPE gains.
  }
  \label{fig:ablationGEO}
  \vspace{-5mm}
\end{figure}

\cref{tab:ablationGEO} and \cref{fig:ablationGEO} examine ego-motion consistency over egocentric rollouts using ATE, RRE, and RPE. Compared with the base variant, enabling EME reduces these ego-motion errors and alleviates accumulated viewpoint drift over time. The full model achieves the best ego-motion metrics, reaching ATE/RRE/RPE of 0.084/5.192/0.021, which indicates closer agreement with the ground-truth viewpoint changes. Qualitatively (\cref{fig:ablationGEO}), EME injection yields more stable viewpoints with less drift across time, and the full model further improves global temporal coherence. Overall, improving ego-motion consistency is important for egocentric rollouts, and our ablation shows that EME is a key factor in reducing viewpoint drift and lowering ATE/RRE/RPE.

\begin{table}[h]
\setlength{\abovecaptionskip}{5pt}
  \vspace{-3mm}
  \caption{\textbf{Ablation on ego-motion consistency.} We evaluate the EME injection variant on long-horizon egocentric rollouts using ATE, RRE, and RPE, and compare it with the base backbone and the full EgoHOI model; $\downarrow$ indicates more consistent ego-motion trajectories and reduced accumulated drift.}
  \label{tab:ablationGEO}
  \centering
  \renewcommand{\arraystretch}{1.15}
  \setlength{\tabcolsep}{4.5pt}
  \resizebox{0.45\linewidth}{!}{ 
    \begin{tabular}{lccc}
      \toprule
      \textbf{Method} 
        & \makecell[c]{ATE} $\downarrow$
        & \makecell[c]{RRE} $\downarrow$
        & \makecell[c]{RPE} $\downarrow$ \\
      \midrule
      Ours (base) & 0.133 & 15.249 & 0.039 \\
      EME Injection & 0.096 & 6.525 & 0.023 \\
      \midrule
      \textbf{Ours (EgoHOI)} & \textbf{0.084} & \textbf{5.192} & \textbf{0.021} \\
      \bottomrule
    \end{tabular}
  }
  \vspace{-3mm}
\end{table}

\subsubsection{Study on Object Integrity.} 
\begin{figure}[h]
\vspace{-3mm}
  \centering
  \includegraphics[width=0.8\textwidth]{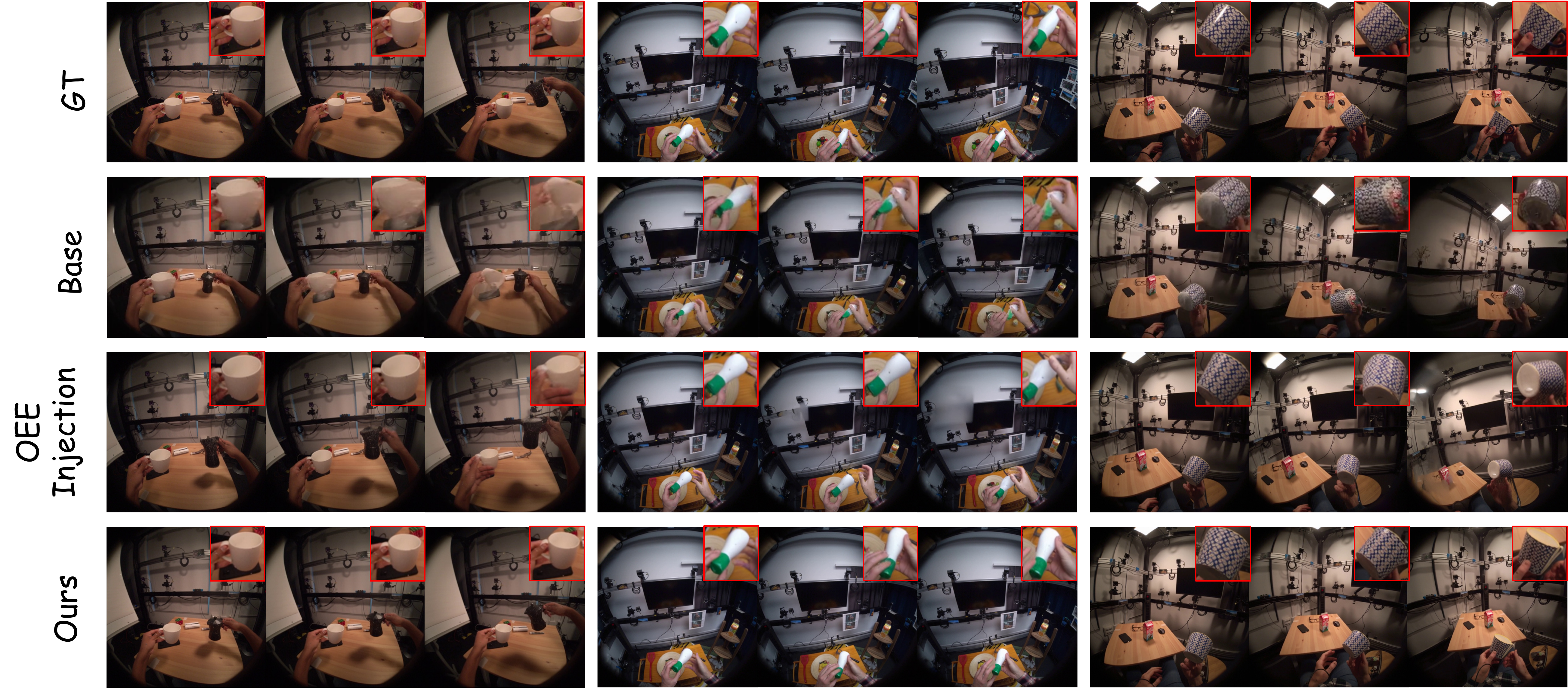} 
  \caption{\textbf{Qualitative results for ablation of object integrity.}
Rows: GT, Base, OEE Injection, Ours (EgoHOI). OEE reduces texture flicker and appearance drift, preserving object-specific details through interaction and occlusion. Zoom-ins highlight more stable patterns and edges. The full model yields the most consistent results.
  }
  \label{fig:APP}
  \vspace{-6mm}
\end{figure}

\cref{tab:ablationAPP} examines the effect of OEE on object integrity over egocentric rollouts. Compared with the base variant, OEE injection improves Object-CLIP from 0.81 to 0.83 and increases VBench Subject Consistency and Imaging Quality from 85.26\% and 52.55\% to 87.90\% and 54.75\%, respectively, indicating more stable object appearance over time. We additionally report object pose errors in this ablation: object position error (OPE) and 2D orientation error (OOE) computed from object masks (details are included in supplementary materials). OEE injection reduces OPE from 0.141 to 0.131 and OOE from 27.739 to 24.124. The full model further improves these metrics to 0.92, 95.51\%, and 64.48\%, while substantially reducing OPE/OOE to 0.015/9.412. Qualitatively (\cref{fig:APP}), OEE injection mitigates drift in object appearance and reduces pose deviations, and the full model yields the most coherent object evolution during periods of frequent hand--object interaction.

\begin{table}[t]
\setlength{\abovecaptionskip}{5pt}
  \caption{\textbf{Ablation on object integrity.} We evaluate the OEE injection variant on egocentric rollouts, where object integrity refers to preserving object identity and maintaining stable object pose under interaction. We report Object CLIP, Subject Consistency, and Imaging Quality ($\uparrow$ is better), together with object position error (OPE) and 2D orientation error (OOE) computed from object masks ($\downarrow$ is better).}
  \label{tab:ablationAPP}
  \centering
  \renewcommand{\arraystretch}{1.15}
  \setlength{\tabcolsep}{4.5pt}
  \resizebox{0.7\linewidth}{!}{ 
    \begin{tabular}{lccccc}
      \toprule
      \textbf{Method} 
        & \makecell[c]{Object\\[-2pt]CLIP} $\uparrow$
        & \makecell[c]{Subject\\[-2pt]Consistency} $\uparrow$
        & \makecell[c]{Imaging\\[-2pt]Quality} $\uparrow$
        & \makecell[c]{OPE} $\downarrow$
        & \makecell[c]{OOE} $\downarrow$\\
      \midrule
      Ours (base) & 0.81 & 85.26\% & 52.55\% & 0.141 & 27.739  \\
      OEE Injection & 0.83 & 87.90\% & 54.75\% & 0.131 & 24.124 \\
      \midrule
      \textbf{Ours (EgoHOI)} & \textbf{0.92} & \textbf{95.51\%} & \textbf{64.48\%} & \textbf{0.015} & \textbf{9.412} \\
      \bottomrule
    \end{tabular}
  }
  \vspace{-3mm}
\end{table}

\section{Conclusion}
We present EgoHOI, an egocentric world model that instantiates action-driven hand–object interaction dynamics with a large video diffusion backbone and a latent internal state. On HOT3D, EgoHOI improves visual quality, ego-motion consistency, and kinematic fidelity. Ablations show that physics-informed embeddings distilled from 3D priors, together with first-frame appearance, preserve object identity, stabilize viewpoints, and maintain realistic interactions.





\bibliographystyle{splncs04}
\bibliography{main}


\appendix




\title{Egocentric World Model for Photorealistic Hand-Object Interaction Synthesis (Supplementary Material)}
\titlerunning{EgoHOI Supplementary Material}
\author{
Dayou Li\inst{1}\textsuperscript{*},
Lulin Liu\inst{1,2}\textsuperscript{*},
Bangya Liu\inst{3},
Shijie Zhou\inst{4},
Jiu Feng\inst{5},
Ziqi Lu\inst{6},
Minghui Zheng\inst{1},
Chenyu You\inst{7},
Zhiwen Fan\inst{1}\textsuperscript{\Letter}
}
\authorrunning{D. Li, L. Liu et al.}

\institute{
$^{1}$Texas A\&M University \quad
$^{2}$University of Minnesota \\
$^{3}$University of Wisconsin--Madison \quad
$^{4}$University of California, Los Angeles \\
$^{5}$University of Texas at Austin \quad
$^{6}$Amazon \\
$^{7}$State University of New York at Stony Brook
}

\maketitle

\begingroup
\renewcommand\thefootnote{*}
\footnotetext{Equal contribution.}
\endgroup

\noindent This supplement is organized as follows:
\begin{itemize}
    \item Section~\ref{sec:motivation} reinforces the motivation of EgoHOI by clarifying how 3D annotations and reconstruction-derived hand, camera, and object signals are used in our framework.
    \item Section~\ref{sec:supp_prior_arch} provides implementation details of the physics-informed embeddings used in EgoHOI;
    \item Section~\ref{suppl:data} describes how the training and test datasets have been built;
    \item Section~\ref{suppl:eval} includes the details of the evaluation metrics, and some definitions of some customized metrics;
    \item Section~\ref{suppl:supplqualitative} contains additional visualization results of the experiments.
    \item Section~\ref{suppl:discussion} discusses the scope of EgoHOI and remaining challenges in evaluating physical plausibility.
\end{itemize}

\section{Motivation}
\label{sec:motivation}
Egocentric Human-Object Interaction (HOI) world models aim to simulate physically grounded interaction dynamics from a first-person perspective driven by user controls. Such predictive rollouts are increasingly important as scalable training and evaluation data for downstream embodied tasks. However, building a reliable egocentric HOI world model is profoundly challenging due to rapid head motions, severe occlusions, and high-DoF hand articulations that abruptly alter contact topologies and object states. Despite recent progress in HOI video generation, this setting remains largely absent in existing pipelines. Current methods are either predominantly studied in static, exocentric views, which do not directly support first-person rollouts with dynamic interactions, or they condition heavily on privileged future object information such as ground-truth trajectories and waypoints, which largely bypasses the core challenge of contact-driven dynamics reasoning. Once these future-state shortcuts are removed, simply scaling RGB data is insufficient for physical accuracy. Instead, the model should be augmented with physics-informed embeddings that inject explicit metric and kinematic structure. Motivated by recent advances in 3D foundation models, we present EgoHOI, an egocentric HOI world model that integrates physics-informed embeddings distilled from 3D estimates into the generative rollout process, providing complementary structure for learning hand kinematics, metric-scale ego motion, and object integrity, while enabling causal simulation of physically plausible hand-object interactions without privileged future object states.


\section{Physics-Informed Embedding Details}
\label{sec:supp_prior_arch}

For completeness, this section provides implementation details for the physics-informed embeddings used in EgoHOI. Following the main paper, we construct three embedding streams from complementary sources of structure: hand kinematics, metric ego motion, and an object-entity anchor tied to the first frame. These embeddings are converted into token sequences aligned with the Wan-DiT latent space and are incorporated into the generative rollout process to regularize action-driven latent dynamics without relying on privileged future object states. Unless otherwise noted, all embedding streams share the token dimension of the Wan-DiT backbone and are aligned to its latent patch layout.

\subsection{Hand Kinematic Embeddings (HKE)}
\label{sec:supp_hke_arch}

HKE captures dense hand motion structure from reconstructed 3D hand estimates, which is essential for contact-consistent HOI rollouts under high-DoF articulation. Rather than encoding hand dynamics as sparse coordinate signals, we use explicit per-frame hand renders obtained from reconstructed meshes as a dense 3D representation and convert them into hand kinematic features aligned with the Wan-DiT latent space.

\subsubsection{Input}
For each frame $t$, we render a hand-kinematics map
$H_t \in \mathbb{R}^{S \times S \times 3}$ from the reconstructed hand mesh,
where $S = 480$.
Stacking an $L$-step rollout yields
$H \in \mathbb{R}^{L \times S \times S \times 3}$.
In our implementation, $L = 81$.
For a batch of size $B$, the temporal hand volume has shape
$B \times L \times S \times S \times 3$.

\subsubsection{Temporal 3D Convolutional Stack}
The temporal hand volume is processed by a temporal 3D convolutional stack with SiLU activations, which gradually increases the channel dimension and maps the hand stream to the shared latent token dimension of the Wan-DiT backbone. The exact layer configuration used in our implementation is summarized in \cref{tab:structHKE3D}.

\begin{table}[h]
\setlength{\abovecaptionskip}{5pt}
  \vspace{-6pt}
  \caption{\textbf{Implementation details of the temporal 3D convolutional stack used in HKE.} For each Conv3d layer we list the kernel size, stride, input$\rightarrow$output channels, and padding.}
  \label{tab:structHKE3D}
  \centering
  \renewcommand{\arraystretch}{1.15}
  \setlength{\tabcolsep}{4.5pt}
  \resizebox{0.6\linewidth}{!}{
    \begin{tabular}{lccccc}
      \toprule
      \makecell[c]{Layer}
        & \makecell[c]{Kernel}
        & \makecell[c]{Stride}
        & \makecell[c]{Channels}
        & \makecell[c]{Padding}\\
      \midrule
      Conv3d & (3,3,3) & (1,1,1) & 3$\rightarrow$16 & (1,1,1) \\
      Conv3d & (3,3,3) & (1,1,1) & 16$\rightarrow$16 & (1,1,1) \\
      Conv3d & (3,3,3) & (1,1,1) & 16$\rightarrow$16 & (1,1,1) \\
      Conv3d & (3,3,3) & (1,2,2) & 16$\rightarrow$16 & (1,1,1) \\
      Conv3d & (3,3,3) & (2,2,2) & 16$\rightarrow$16 & (1,1,1) \\
      Conv3d & (3,3,3) & (2,2,2) & 16$\rightarrow$16 & (1,1,1) \\
      Conv3d & (1,2,2) & (1,2,2) & 16$\rightarrow$5120 & (0,0,0) \\
      \bottomrule
    \end{tabular}
  }
  \vspace{-6pt}
\end{table}

\subsubsection{Reference-Hand Branch}
To stabilize hand identity and reduce temporal drift, we add a reference-hand branch that encodes the first-frame hand render into a compact feature through a 2D convolutional pyramid. This branch produces a first-frame hand feature that is incorporated into the initial latent during latent-space integration. The exact layer configuration is summarized in \cref{tab:structHKEref}.

\begin{table}[h]
\setlength{\abovecaptionskip}{5pt}
  \vspace{-6pt}
  \caption{\textbf{Implementation details of the reference-hand branch used in HKE.} This branch processes the first-frame hand render; for each Conv2d layer we list the kernel size, stride, input$\rightarrow$output channels, and padding.}
  \label{tab:structHKEref}
  \centering
  \renewcommand{\arraystretch}{1.15}
  \setlength{\tabcolsep}{4.5pt}
  \resizebox{0.6\linewidth}{!}{
    \begin{tabular}{lccccc}
      \toprule
      \makecell[c]{Layer}
        & \makecell[c]{Kernel}
        & \makecell[c]{Stride}
        & \makecell[c]{Channels}
        & \makecell[c]{Padding}\\
      \midrule
      Conv2d & 3 & 1 & 3$\rightarrow$16 & 1 \\
      Conv2d & 3 & 1 & 16$\rightarrow$16 & 1 \\
      Conv2d & 3 & 1 & 16$\rightarrow$16 & 1 \\
      Conv2d & 3 & 2 & 16$\rightarrow$16 & 1 \\
      Conv2d & 3 & 2 & 16$\rightarrow$16 & 1 \\
      Conv2d & 3 & 2 & 16$\rightarrow$20 & 1 \\
      \bottomrule
    \end{tabular}
  }
  \vspace{-6pt}
\end{table}

\subsection{Ego-Motion Embeddings (EME)}
\label{sec:supp_eme_arch}

EME encodes metric head trajectory estimates into embeddings aligned with the latent space, enabling geometry-consistent viewpoint evolution over time. Following the main paper, we represent ego motion as an image-aligned Pl\"ucker ray field and map the resulting trajectory representation into ego-motion tokens compatible with the Wan-DiT backbone.

\subsubsection{Input and Pl\"ucker Ray Field}
For each frame, the calibrated intrinsics $K_t$ and extrinsics
$E_t = [R_t \mid t_t]$, measured relative to the first frame, are converted into an image-aligned Pl\"ucker ray field on a regular grid.
For each pixel $(u,v)$, we form a 6D ray representation
$p_{u,v} = (o \times d_{u,v},\, d_{u,v})$,
where $o$ is the head-frame optical center and:
\begin{equation}
    d_{u,v} = R_t K_t^{-1}
\begin{bmatrix}
u\\v\\1
\end{bmatrix}
+ t_t
\end{equation}

Stacking the per-frame Pl\"ucker tensors over time yields a trajectory volume of shape
$P \in \mathbb{R}^{B \times 6 \times L \times S \times S}$,
where $L = 81$ and $S = 480$ in our implementation.

\subsubsection{Temporal Causal Downsampling}
We first process the trajectory volume with a temporal causal downsampler composed of three causal 3D convolutional blocks. Each block uses kernel size $3 \times 3 \times 3$, GroupNorm, and SiLU activation. The first two blocks use stride $(2,2,2)$, and the third uses stride $(1,2,2)$. This stage reduces the temporal and spatial resolutions while preserving metric structure under head motion, and produces a feature map
$F_0 \in \mathbb{R}^{B \times 64 \times 21 \times 60 \times 60}$.

\subsubsection{Hybrid Spatiotemporal Processing}
After temporal downsampling, the feature map is processed by a stack of hybrid spatiotemporal stages that combine 2D residual convolutions with temporal self-attention. In our implementation, the stage channels are $[64,128,256]$. Each stage reshapes the tensor from $[B,C,21,H,W]$ to $[B \cdot 21, C, H, W]$, applies two $3 \times 3$ Conv2d layers with stride $1$, and uses an additional stride-$2$ Conv2d in stages that reduce the spatial resolution. The features are then reshaped back to $[B,C_i,21,H_i,W_i]$ and passed through a temporal self-attention block along the 21-frame dimension. This design preserves spatially resolved constraints on viewpoint evolution instead of collapsing ego motion into a low-dimensional pose vector.

\subsubsection{Projection Head and Patchification}
The final feature volume is normalized and projected before tokenization. The output head consists of: (i) a GroupNorm layer, (ii) a $1 \times 1 \times 1$ convolution for channel mixing, and (iii) a 3D convolutional patchifier with kernel size $(1,2,2)$ and the same stride. After patchification, the feature map has shape
(B, 5120, 21, 30, 30).
Flattening the $(21,30,30)$ grid into the sequence dimension yields ego-motion tokens
$T_{\mathrm{EME}} \in \mathbb{R}^{B \times N_{\mathrm{EME}} \times d}$,
where $N_{\mathrm{EME}} = 21 \times 30 \times 30$ and $d = 5120$.
These tokens are later injected into the latent dynamics through the zero-initialized linear adapters described in the main paper.

\subsection{Object-Entity Embeddings (OEE)}
\label{sec:supp_oee_arch}

OEE anchors the manipulated object to the first frame to provide a persistent object-centric reference throughout the rollout. Concretely, we use the first-frame object segmentation image, obtained from an off-the-shelf segmentation model, to isolate object-centric evidence and reduce background interference under viewpoint changes. The segmentation is encoded by the frozen video VAE encoder into a latent grid, and a 3D convolutional patchifier converts it into tokens aligned with the main latent stream.

\subsubsection{Input and VAE Latent}
From the first frame, we obtain an object segmentation image and resize it to $480 \times 480$.
The image is normalized to $[-1,1]$, arranged as $[B,3,H,W]$, and expanded to $[B,3,1,H,W]$ before being passed to the frozen Wan VAE encoder.
Operating in the same latent parameterization as the backbone, OEE yields a latent grid of shape
[B,16,21,60,60].

\subsubsection{Object Patchifier}
The latent grid is fed directly into a 3D convolutional patchifier without further convolutional processing beforehand.
The patchifier uses kernel size $(1,2,2)$ and the same stride, preserving the temporal length of 21 while reducing the spatial grid from $60 \times 60$ to $30 \times 30$.
The output channel dimension equals the Wan-DiT hidden dimension, yielding a feature map of [B,5120,21,30,30].

\subsubsection{Object-Entity Tokens and Positional Shift}
Flattening the $(21,30,30)$ grid into the sequence dimension yields object-entity tokens
$T_{\mathrm{OEE}} \in \mathbb{R}^{B \times N_{\mathrm{OEE}} \times d}$,
where $d = 5120$ matches the hidden dimension of the Wan-DiT token space.
To spatially anchor these tokens as an independent reference stream, we assign shifted rotary position indices when constructing the RoPE frequency grid.
During latent-space integration, the OEE tokens are appended to the active token sequence and participate as KV-only context in self-attention, while only the main tokens are decoded.
This design provides a persistent entity reference that reduces object drift in long rollouts.

\section{Data Preparation}
\label{suppl:data}
\subsection{HOT3D Benchmark}
HOT3D-Clips is a curated subset of the HOT3D dataset, released in WebDataset format as synchronized frame streams instead of ready-made videos. Each clip contains 150 egocentric frames of hand–object interaction with metric 3D annotations, including calibrated camera intrinsics and extrinsics, a world-aligned camera trajectory, articulated 3D hand kinematics, and 6-DoF object pose linked to scanned meshes. Project Aria provides multiple synchronized camera streams per frame; EgoHOI uses the forward-facing RGB stream (stream id is 214-1, labeled camera-rgb) as visual input, together with the associated 3D priors. Frames from stream 214-1 are grouped into 150-frame clips at 16 FPS. Clips are then randomly partitioned at the clip level, with 90\% used for training and 10\% reserved for testing, so that no frame from a test clip appears in training. In total, we have 1,364 samples for training and 152 samples for testing.

\begin{figure}[h]
  \centering
  \includegraphics[width=0.6\textwidth]{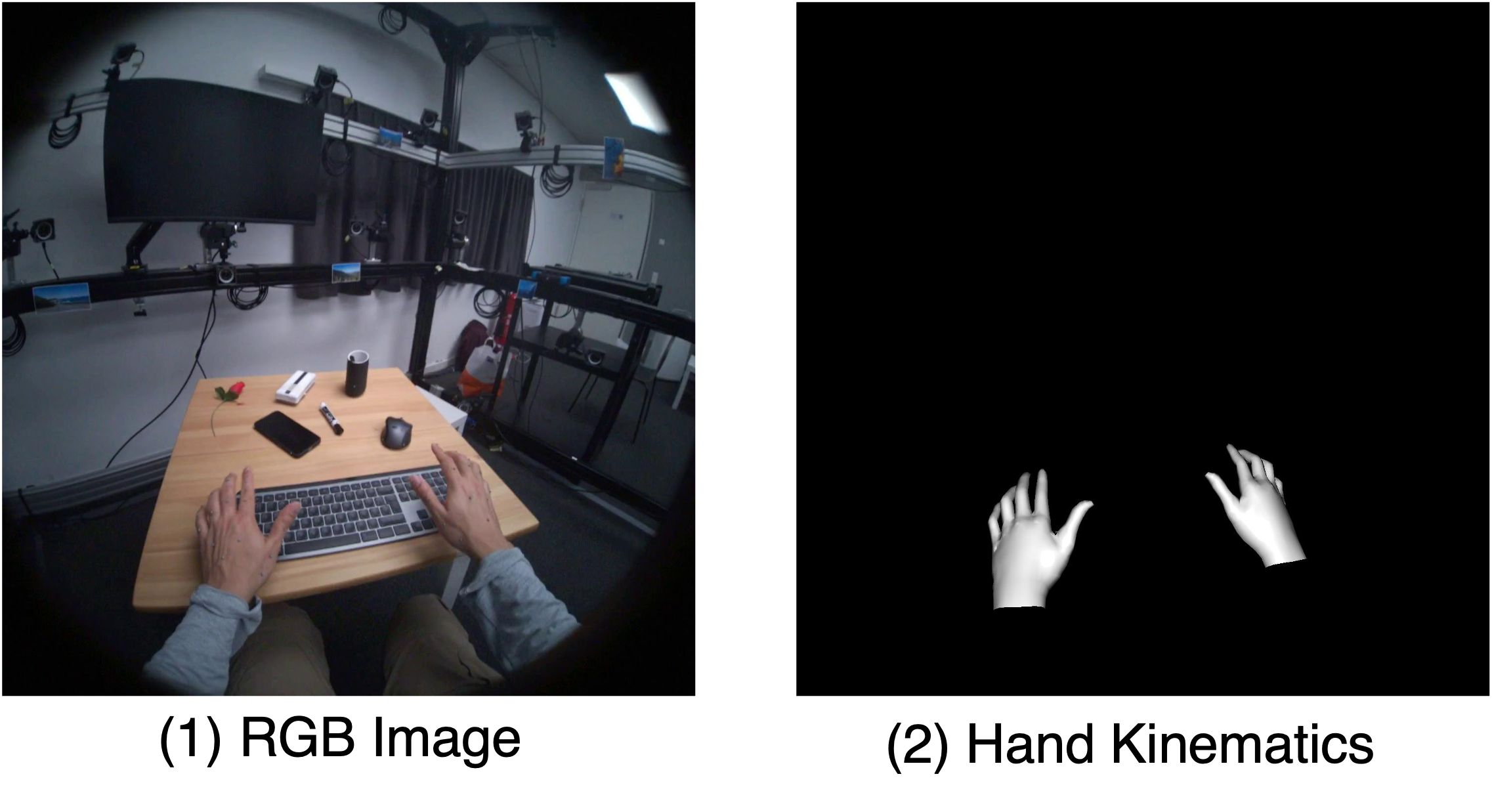} 
  \caption{\textbf{Visualization of hand kinematics.} For each frame, we use the egocentric RGB image (left) together with a rendered
hand-kinematics map (right) that visualizes the articulated meshes of both hands in the camera coordinate frame.
  }
  \label{fig:visualhand}
\end{figure}

In addition, we convert the 3D hand annotations in HOT3D-Clips into image-space renderings that serve as input to the kinematics stream. For each frame, the MANO-based hand meshes are projected with the calibrated intrinsics and extrinsics of stream 214-1 and rasterized onto a blank canvas, producing a grayscale rendering of the visible hands in the current viewpoint, aligned pixel-wise with the corresponding RGB frame (as shown in \cref{fig:visualhand}). These hand-only renderings preserve the precise 3D articulation while stripping away appearance, which makes the kinematic prior independent of texture and lighting.

Since the Wan-based backbone in EgoHOI can only take 81 frames in a single rollout, a full 150-frame clip cannot be processed in one pass. During training, 81-frame trajectories are therefore sampled on the fly by sliding a temporal window over each clip with a stride of 5 frames. For every window, the corresponding RGB frames from stream 214-1 and their 3D priors are collected. These overlapping windows share the same scene but have different starting times, which increases trajectory diversity.

\subsection{Text Prompt Generation}
\label{sec:supp_prompts}
To train text-conditioned baselines, each HOT3D clip is paired with an automatic caption generated by Tarsier2-7B~\cite{yuan2025tarsier2}, a large video-language model designed for detailed, low-hallucination video description and comprehensive video understanding. Tarsier2 combines a vision encoder, a lightweight adaptor, and a language model, and is trained in multiple stages on large-scale video–text corpora, which yields accurate and temporally grounded captions. For each 150-frame clip, we uniformly sample representative keyframes and feed them to Tarsier2 together with an instruction asking for a concise yet detailed description of the scene and hand–object interaction. The resulting caption is stored as the text prompt for that clip and reused for all sliding-window segments derived from it when training the text-conditioned baseline world models.

\section{Evaluation Metrics}
\label{suppl:eval}
The evaluation protocol examines generated rollouts from four complementary perspectives: visual prediction, ego-trajectory consistency, object integrity, and kinematic fidelity. Visual prediction is assessed using standard image and video quality metrics, together with VBench scores that reflect perceptual and temporal quality. Ego-trajectory consistency is quantified from estimated camera trajectories. Object integrity is evaluated using Object-CLIP, object position error (OPE), and object orientation error (OOE), which together measure object appearance preservation, location preservation, and in-plane orientation preservation under interaction. Kinematic fidelity is measured from hand reconstructions obtained from both ground-truth and generated videos. Overall, the evaluation protocol combines standard metrics for comparison with prior work and custom measures designed for egocentric hand-object interaction.

\subsection{Visual Prediction Metrics}
Visual prediction is evaluated using standard image quality metrics, including peak signal-to-noise ratio (PSNR), structural similarity (SSIM), perceptual similarity (LPIPS), and Object-CLIP. These metrics capture complementary aspects of rollout quality, ranging from pixel-level fidelity and perceptual similarity to object appearance preservation between ground-truth frames and the corresponding generated frames. Since these measures are widely used in video generation and reconstruction, they also enable direct comparison with prior world models.

\subsection{Ego-Motion Consistency Metrics}
To quantify ego-motion consistency, we evaluate camera trajectories using Absolute Trajectory Error (ATE), Relative Pose Error (RPE), and Relative Rotation Error (RRE). Concretely, we apply MapAnything~\cite{keetha2025mapanything} to both the ground-truth video and the generated rollout under the same configuration to obtain two sequences of estimated head poses. These pose sequences are then temporally synchronized and aligned in a common coordinate frame, after which ATE, RPE, and RRE are computed in the standard way. This protocol measures whether the generated rollout preserves ego-motion that is geometrically consistent with the reference trajectory rather than merely producing visually plausible frames.

\subsection{Object-CLIP Score}
We evaluate object appearance preservation by computing CLIP similarity between masked ground-truth frames and the corresponding generated frames. For each evaluation clip, we load object-masked RGB frames, align the ground-truth and generated sequences in time, and resize the ground-truth frames to match the generated resolution for one-to-one frame pairing. All frames are encoded with a single CLIP image encoder implemented with the \texttt{open\_clip} backend using ViT-B/32 pretrained on LAION-2B S/34B, with images resized and normalized to $224 \times 224$. We extract L2-normalized embeddings and compute cosine similarity for each paired frame $(I_t^{\mathrm{gt}}, I_t^{\mathrm{gen}})$. The Object-CLIP score for one clip is the average similarity over all paired frames, and the reported value is the mean over the evaluation set.

\subsection{Object Position and Orientation Metrics}
\label{sec:object_pose_metrics}

To provide a more fine-grained evaluation of object integrity, we introduce two object-level metrics computed from object masks in the ground-truth and generated frames: Object Position Error (OPE) and Object Orientation Error (OOE). These metrics complement Object-CLIP by explicitly measuring geometric consistency in image space.

\subsubsection{Object Position Error (OPE)}
For each frame, we extract binary object masks from the ground-truth and generated images. Given a mask with foreground pixels $\{(x_i,y_i)\}_{i=1}^{A}$, its centroid is:
\begin{equation}
c_x=\frac{1}{A}\sum_{i=1}^{A}x_i,
\qquad
c_y=\frac{1}{A}\sum_{i=1}^{A}y_i.
\end{equation}
Let $(c_x^{\mathrm{gt}}, c_y^{\mathrm{gt}})$ and $(c_x^{\mathrm{gen}}, c_y^{\mathrm{gen}})$ denote the centroids of the ground-truth and generated masks. We define the frame-level position error as the centroid distance normalized by the image diagonal:
\begin{equation}
e_{\mathrm{pos},t}
=
\frac{
\sqrt{
(c_x^{\mathrm{gt}}-c_x^{\mathrm{gen}})^2+
(c_y^{\mathrm{gt}}-c_y^{\mathrm{gen}})^2
}
}{
\sqrt{H^2+W^2}
}.
\end{equation}
If both masks are empty, we set $e_{\mathrm{pos},t}=0$; if only one mask is empty, we assign a penalty of $1.0$.

\subsubsection{Object Orientation Error (OOE)}
For a non-empty mask, we estimate its in-plane orientation from the second-order central moments:
\begin{align}
\mu_{20} &= \frac{1}{A}\sum_{i=1}^{A}(x_i-c_x)^2, \\
\mu_{02} &= \frac{1}{A}\sum_{i=1}^{A}(y_i-c_y)^2, \\
\mu_{11} &= \frac{1}{A}\sum_{i=1}^{A}(x_i-c_x)(y_i-c_y),
\end{align}
which define the principal-axis angle:
\begin{equation}
\theta=\frac{1}{2}\operatorname{atan2}(2\mu_{11}, \mu_{20}-\mu_{02}).
\end{equation}
To discard nearly isotropic masks, we compute the anisotropy score:
\begin{equation}
\alpha=\frac{\lambda_1-\lambda_2}{\lambda_1+\lambda_2+\varepsilon},
\qquad \varepsilon=10^{-12},
\end{equation}
where $\lambda_1,\lambda_2$ are the covariance eigenvalues. Orientation is considered valid only when $A \ge 20$ and $\alpha \ge 0.15$. For valid ground-truth and generated masks, the angular difference is computed modulo $\pi$ and folded into $[0,\pi/2]$:
\begin{equation}
\Delta\theta_t
=
\min\!\left(
\left|\theta_t^{\mathrm{gt}}-\theta_t^{\mathrm{gen}}\right| \bmod \pi,\,
\pi-\left(\left|\theta_t^{\mathrm{gt}}-\theta_t^{\mathrm{gen}}\right| \bmod \pi\right)
\right).
\end{equation}
The frame-level orientation error is:
\begin{equation}
e_{\mathrm{ori},t}=\Delta\theta_t \cdot \frac{180}{\pi}.
\end{equation}
Frames with invalid orientation are excluded from OOE. For each clip, OPE is averaged over all frames, and OOE is averaged over frames with valid orientation. We report the dataset-level macro-average of both metrics, together with the valid-frame ratio used for OOE.
\begin{table}[t]
\setlength{\abovecaptionskip}{5pt}
  \caption{\textbf{Ablation on object integrity.} In addition to the object integrity analysis in the main paper, we report object position error (OPE) and 2D orientation error (OOE) to further quantify how well each method preserves object location and in-plane orientation under interaction. $\downarrow$ is better for both metrics.}
  \label{tab:ablationAPP}
  \centering
  \renewcommand{\arraystretch}{1.15}
  \setlength{\tabcolsep}{4.5pt}
  \resizebox{0.4\linewidth}{!}{ 
    \begin{tabular}{lcc}
      \toprule
      \textbf{Method} 
        & \makecell[c]{OPE} $\downarrow$
        & \makecell[c]{OOE} $\downarrow$\\
      \midrule
      Wan & 0.078 & 19.295  \\
      Cosmos 2B & 0.112 & 23.823 \\
      Cosmos 14B & 0.108 & 23.312 \\
      Uni3C & 0.083 & 24.751 \\
      \midrule
      \textbf{Ours (EgoHOI)} & \textbf{0.015} & \textbf{9.412} \\
      \bottomrule
    \end{tabular}
  }
  \vspace{-3mm}
\end{table}


\subsection{Kinematic Fidelity Metrics}
We evaluate kinematic fidelity by applying HaMeR~\cite{pavlakos2024reconstructing} to both ground-truth and generated frames. The evaluation reports three metrics: the missing ratio (MR), which measures how often a valid hand reconstruction is absent when a ground-truth hand is present; the mean per-joint position error (MPJPE), which measures 3D joint accuracy; and a root mean squared error (RMSE) computed on hand segmentation images. MPJPE is a standard metric in 3D hand pose estimation, while MR and segmentation RMSE capture characteristic failure modes in egocentric HOI rollouts, including complete hand disappearance and gradual degradation of the visible arm extent in image space.

\subsection{Hand Visibility Missing Ratio (MR)}
The missing ratio measures how often the model fails to provide a valid hand prediction when a ground-truth hand is present. Let $\mathcal{H}$ denote the set of all ground-truth hand instances indexed by time and hand identity:
\begin{equation}
    \mathcal{H} = \{(t, i)\mid \text{hand } i \text{ is annotated at frame } t\}
\end{equation}
For each ground-truth instance $(t, i)$, define an indicator:
\begin{equation}
    v_{t,i} = 
    \begin{cases}
        1, & \text{if HaMeR reconstructs a valid hand} \\ & \text{for instance } (t, i),\\[2pt]
        0, & \text{otherwise},
    \end{cases}
\end{equation}

where validity is determined by the presence of a confident reconstruction and a bounded MPJPE. The missing ratio is then defined:
\begin{equation}
    \mathrm{MR} 
    = 1 - \frac{1}{|\mathcal{H}|} \sum_{(t,i)\in\mathcal{H}} v_{t,i}
\end{equation}
A lower $\mathrm{MR}$ indicates that hands are reconstructed more reliably and remain present throughout the rollout, which is crucial for long-horizon egocentric sequences where hands frequently enter and leave the field of view.

\subsection{RMSE on Hand Segmentation Images}
\begin{figure}[h]
  \centering
  \includegraphics[width=0.4\textwidth]{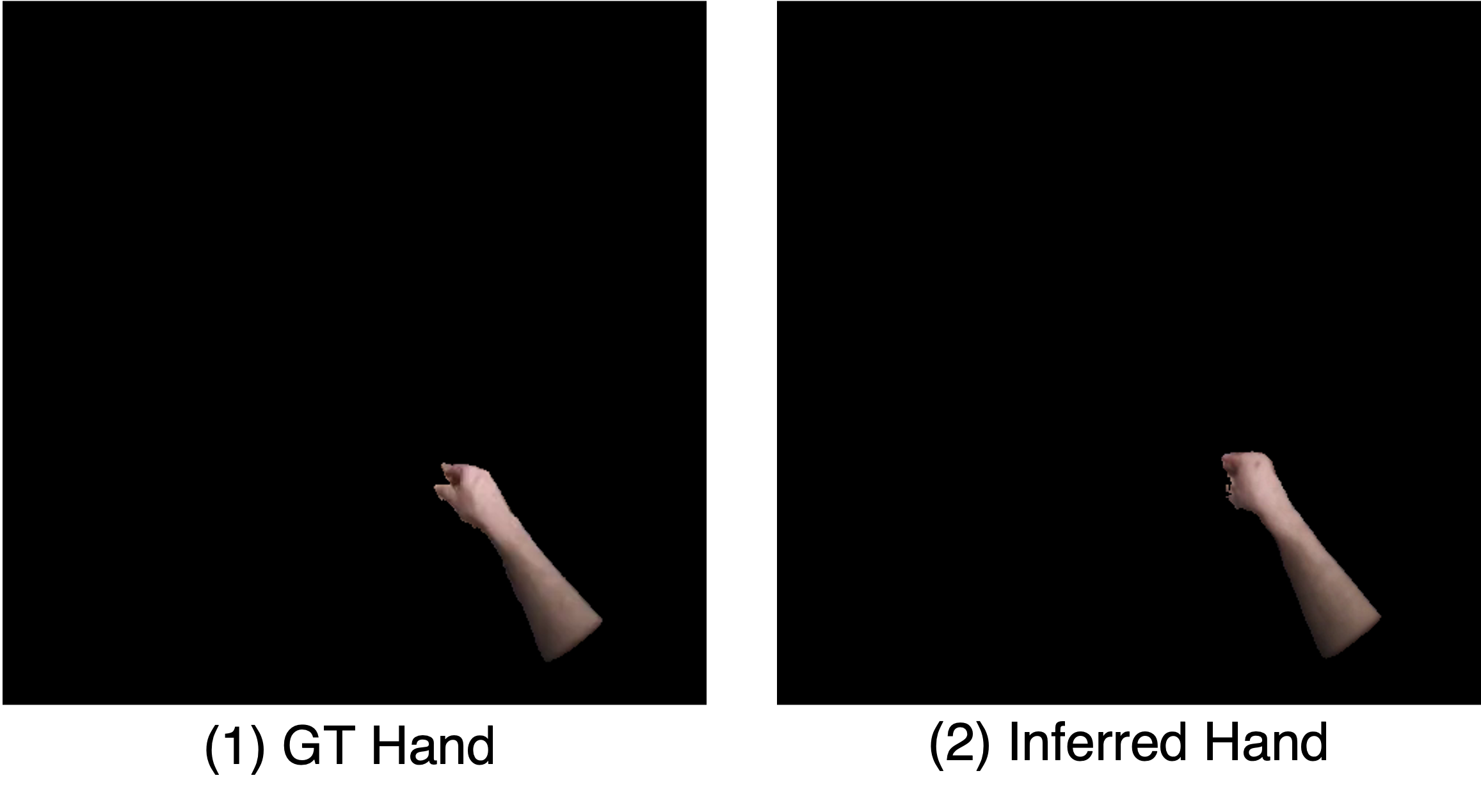} 
  \caption{\textbf{Hand Segmentation for RMSE Evaluation.} For each frame, we extract the ground-truth hand region (left) and the
corresponding inferred hand region produced by EgoHOI (right). The RMSE is computed as the per-pixel root-mean-square error between these two hand masks, averaged over all frames.
  }
  \label{fig:visualhandseg}
\end{figure}

To directly assess how well the generated video preserves the spatial extent of the visible arm, a root mean squared error is computed between hand segmentation images extracted from the ground-truth and generated frames. For each frame $t$, the Segment Anything~\cite{kirillov2023segment} model is applied to obtain an arm mask from both the ground-truth frame and the corresponding generated frame under the same prompt. This yields two segmentation images (as shown in \cref{fig:visualhandseg}, represented as binary or probabilistic masks $M_t(x)$ and $\hat{M}_t(x)$ for each pixel $x$ in the image domain $\Omega$, where $M_t(x) \in \{0,1\}$ denotes the ground-truth arm mask and $\hat{M}_t(x) \in [0,1]$ denotes the predicted arm probability.
With $T$ frames and pixel domain $\Omega$, the RMSE on hand segmentation images is defined as:
\[
    \mathrm{RMSE}_{\mathrm{seg}}
    = \sqrt{\frac{1}{T\,|\Omega|} 
      \sum_{t=1}^{T} \sum_{x \in \Omega} 
      \bigl( M_t(x) - \hat{M}_t(x) \bigr)^{2} }.
\]
A lower $\mathrm{RMSE}_{\mathrm{seg}}$ indicates a closer match between the generated and ground-truth arm regions in pixel space, penalizing errors in arm shape, position, and scale. This metric focuses on the spatial footprint of the arm rather than joint coordinates, and it complements MPJPE and MR by capturing subtle drifts in arm appearance that may not be reflected in joint-level errors alone.
Taken together, MR and the segmentation RMSE isolate two characteristic failure modes in egocentric HOI rollouts, namely complete disappearance of the hand and gradual degradation of arm extent in image space, while standard frame and camera metrics account for global appearance and viewpoint consistency.

\subsection{VBench Scores}
VBench~\cite{zheng2025vbench} provides metrics that decompose video generation quality into temporal consistency and per-frame visual quality. In the egocentric HOI setting on HOT3D, only a subset of VBench dimensions can be applied directly to our generated rollouts without requiring text prompts or class labels. Therefore, the evaluation reports six scores: Subject Consistency, Background Consistency, Motion Smoothness, Dynamic Degree, Aesthetic Quality, and Imaging Quality. Other VBench dimensions that require explicit category, attribute, or style specifications are not used in this benchmark.

Subject Consistency and Background Consistency measure how stable the foreground entities and the surrounding scene remain over time. High Subject Consistency indicates that the hand and manipulated object preserve identity and appearance across the rollout, rather than gradually drifting to different shapes or textures. High Background Consistency indicates that walls, tables, and other scene elements maintain coherent geometry and semantics under ego-motion, rather than exhibiting flickering layouts or implausible structural changes. Motion Smoothness uses a frame-interpolation prior to assess whether motion between adjacent frames evolves gradually and remains temporally coherent, without abrupt changes.

Dynamic Degree aggregates optical flow magnitude to characterize the overall level of motion in a video. This metric reflects how active the sequence is, rather than how physically correct the dynamics are. In HOT3D, ground-truth clips already exhibit moderate ego-camera motion and purposeful hand-object interaction, so Dynamic Degree should be interpreted together with the consistency metrics. Very low values correspond to nearly static views or frozen hands, which under-represent the intended trajectories. Extremely high values indicate unnecessarily strong or unstable motion that no longer matches the underlying egocentric path. An appropriate Dynamic Degree in this benchmark therefore reflects a balanced amount of movement, rather than being interpreted as ``the higher the better.''

Aesthetic Quality and Imaging Quality capture complementary aspects of frame-level appearance. Aesthetic Quality correlates with judgments of composition, color, and overall visual appeal, while Imaging Quality focuses on low-level fidelity, including blur, noise, and exposure. Reporting these scores together verifies that improvements in subject and background consistency or motion stability do not come at the expense of perceived visual quality.
\begin{figure*}[h]
  \centering
  \includegraphics[width=\textwidth]{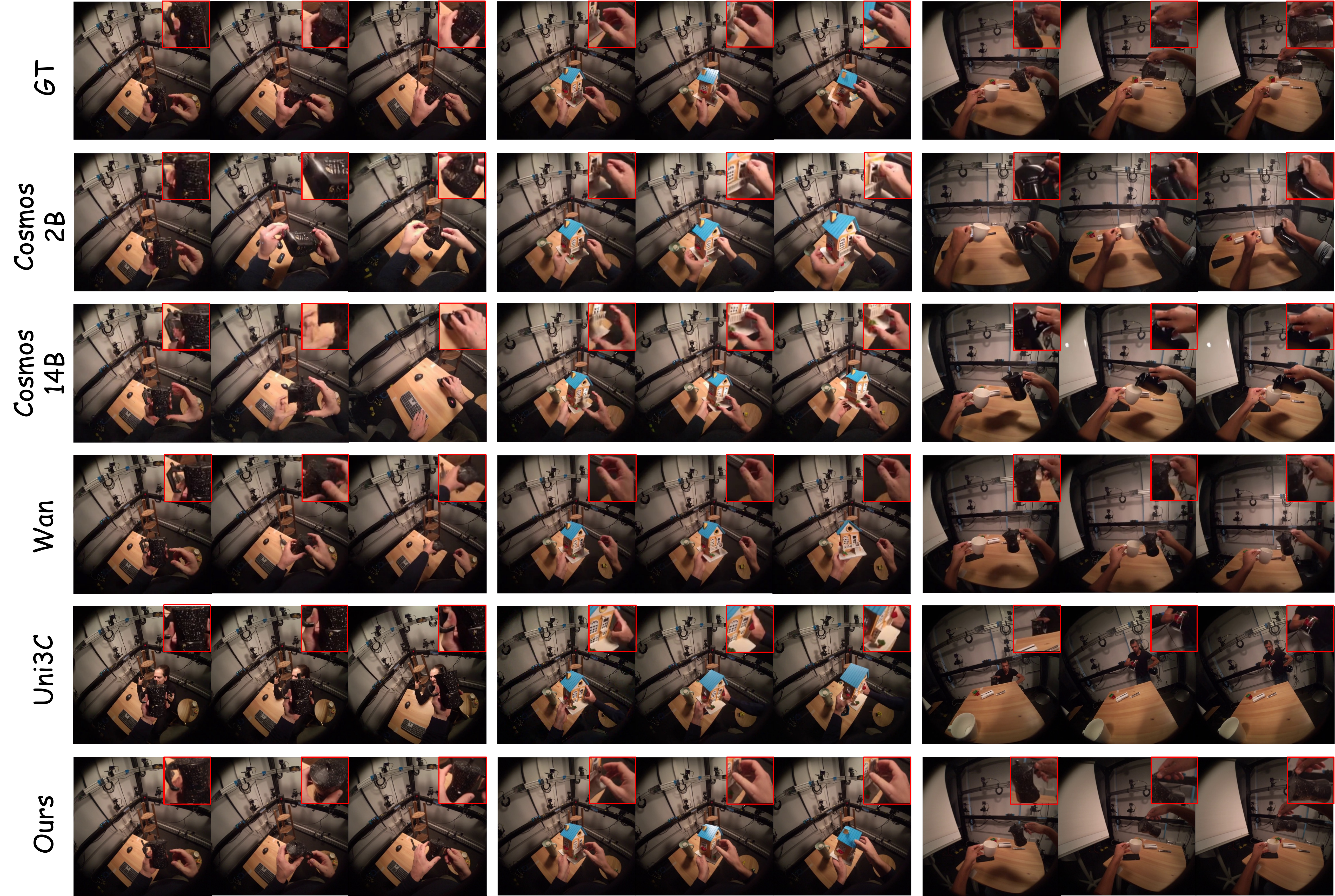} 
  \caption{\textbf{Qualitative comparison with baselines (part I).} Rows from top to bottom show the ground truth (GT), Cosmos-2B, Cosmos-14B, Wan, Uni3C, and our EgoHOI model. Each column block shows consecutive frames from one egocentric test sequence in temporal order. All methods receive the same first frame as input. Compared with the baselines, our model better preserves hand and object geometry, maintains object identity, and produces more stable interaction dynamics over time. See the zooming boxes for comparison of the fine-grained details.
  }
  \label{fig:qual2}
\end{figure*}

\begin{figure*}[h]
  \centering
  \includegraphics[width=\textwidth]{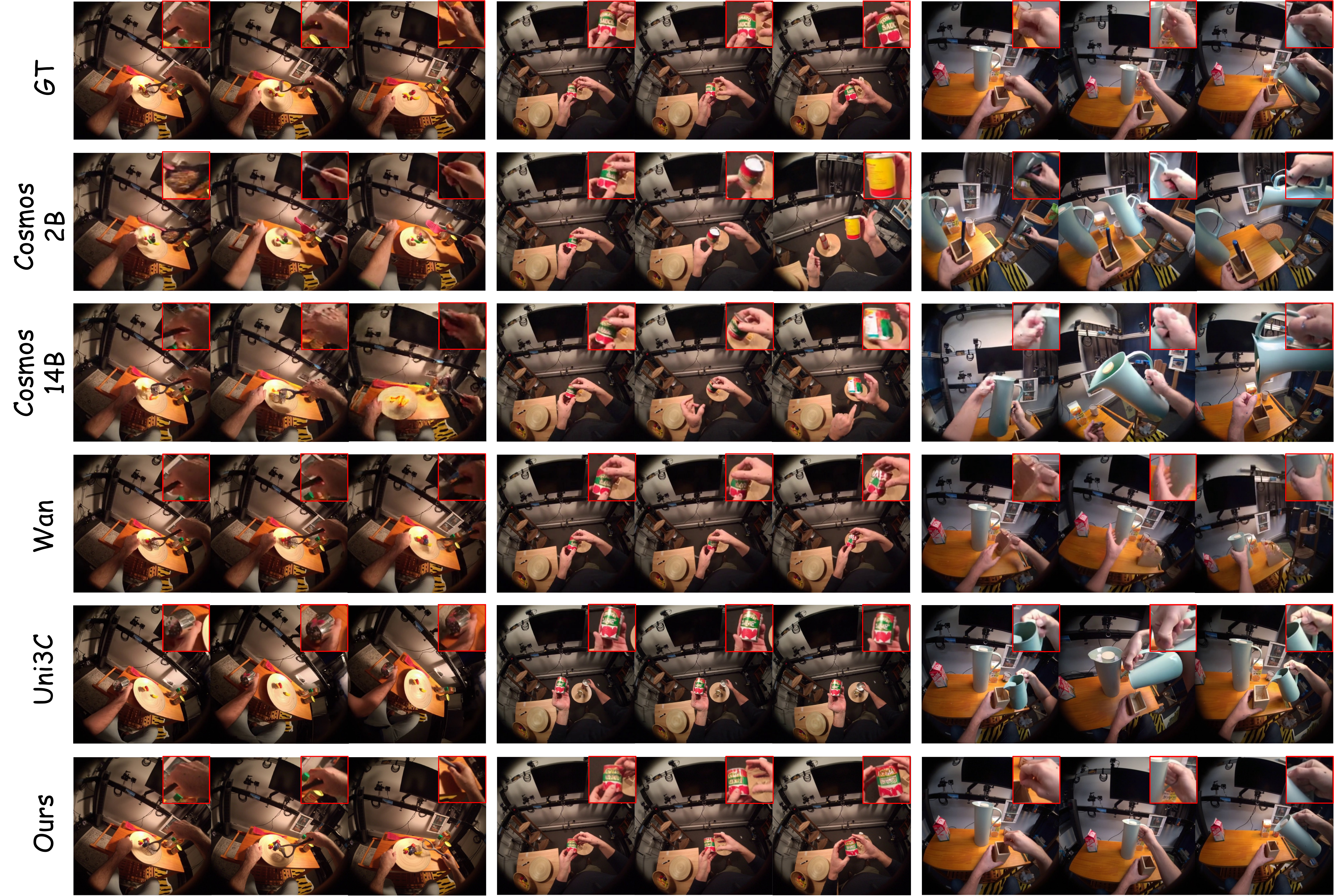} 
  \caption{\textbf{Qualitative comparison with baselines (part II).} Rows from top to bottom show the ground truth (GT), Cosmos-2B, Cosmos-14B, Wan, Uni3C, and our EgoHOI model. Each column block shows consecutive frames from one egocentric test sequence in temporal order. All methods receive the same first frame as input. Compared with the baselines, our model better preserves hand and object geometry, maintains object identity, and produces more stable interaction dynamics over time. See the zooming boxes for comparison of the fine-grained details.}
  \label{fig:qual1}
\end{figure*}

\begin{figure}[t]
  \centering
  \includegraphics[width=1\textwidth]{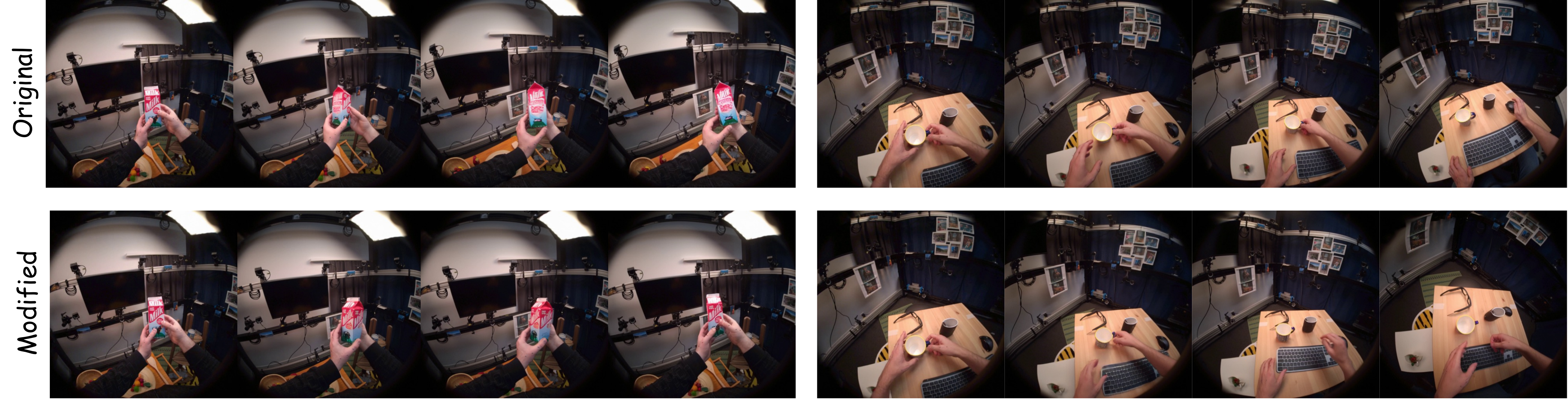} 
  \caption{\textbf{Qualitative analysis under different user inputs.} For each example, both rows start from the same first frame but use different user inputs. EgoHOI produces distinct rollout outcomes with different viewpoint evolution and interaction trajectories, while preserving coherent scene structure and plausible hand-object interaction.}
  \label{fig:alt}
\end{figure}
\section{Qualitative Results}
\label{suppl:supplqualitative}

In this section, we provide additional qualitative results that complement the main paper. We first compare EgoHOI with baseline methods on representative HOT3D examples, highlighting improvements in hand geometry, contact realism, and local appearance around the interaction region. We then provide qualitative examples under different user inputs to illustrate how the generated rollout varies with changes in ego-motion input.

\subsection{Comparison with Baselines}

As shown in Fig.~\ref{fig:qual2} and Fig.~\ref{fig:qual1}, our method produces images that are visually more realistic and spatially more coherent: hand shapes follow plausible anatomy, contact regions agree with the intended manipulation, and local appearance around the hands preserves sharper details and cleaner boundaries. Competing models frequently exhibit deformed or incomplete fingers, inaccurate hand-object overlap, and artifacts near the contact area, which reduces the perceived realism of the interaction. These qualitative comparisons indicate that incorporating our 3D priors yields more reliable hand geometry and higher-fidelity hand appearance in egocentric views.
\uline{Additional visualization videos are available on the webpage attached to the supplementary materials.}

\subsection{Qualitative Analysis under Different User Inputs}

To further illustrate the action-driven world modeling behavior of EgoHOI, we provide qualitative examples generated from the same first frame under different user inputs. As shown in Fig.~\ref{fig:alt}, changing the user input leads to distinct rollout outcomes, while the generated videos remain physically plausible and preserve coherent hand-object interaction dynamics over time. These examples suggest that EgoHOI does not merely reproduce visually plausible continuations from a fixed initial observation; instead, it responds to different user controls by simulating different and plausible interactions.

\section{Discussion}
\label{suppl:discussion}

Egocentric HOI world models are critical for simulating physically grounded first-person rollouts. In this work, EgoHOI breaks away from privileged future-state shortcuts by integrating physics-informed embeddings into the generative rollout process, enabling photorealistic, contact-consistent hand-object interactions from action signals alone. The empirical gains in visual prediction, ego-trajectory consistency, kinematic fidelity, and object integrity suggest that explicit metric and kinematic structure provides an effective inductive bias for egocentric HOI simulation. At the same time, fully evaluating physical plausibility remains an open problem. While our current benchmark captures multiple complementary aspects of rollout quality through reconstruction-derived and perceptual metrics, future work should develop more direct contact-aware and dynamics-aware evaluations for first-person hand-object interaction rollouts.






\end{document}